\documentclass[journal,compsoc]{IEEEtran}

\ifCLASSOPTIONcompsoc
\usepackage[nocompress]{cite}
\else
\usepackage{cite}
\fi

\ifCLASSINFOpdf
\usepackage[pdftex]{graphicx}
\else
\usepackage[dvips]{graphicx}
\DeclareGraphicsExtensions{.eps}
\fi
\usepackage{booktabs}
\usepackage{bbm}
\usepackage{threeparttable}
\usepackage{color,graphicx}
\usepackage{amsmath}
\usepackage{amsthm}
\usepackage[ruled]{algorithm2e}
\usepackage{bm}
\usepackage{multirow}
\usepackage{booktabs}
\usepackage{amssymb}

\usepackage{bm}
\usepackage{multirow}
\usepackage{booktabs}
\usepackage{amssymb}
\newtheorem{definition}{Definition}
\newtheorem{proposition}{Proposition}

\usepackage{changes}
\usepackage{subfigure}
\definechangesauthor[name={Review 1}, color=orange]{R1}
\definechangesauthor[name={Review 2}, color=blue]{R2}
\definechangesauthor[name={Review 1,2}, color=red]{R1+R2}

\usepackage{verbatim}

\providecommand{\zkreffig}[1]{Fig.~\ref{#1}}

\ifCLASSOPTIONcompsoc
\usepackage[caption=false,font=footnotesize,labelfont=sf,textfont=sf]{subfig}
\else
\usepackage[caption=false,font=footnotesize]{subfig}
\fi

\bstctlcite{IEEEexample:BSTcontrol}

\begin{document}
	\title{Deep Compatible Learning for Partially-Supervised Medical Image Segmentation}

	\author{Ke~Zhang and~Xiahai Zhuang$^*$
		\IEEEcompsocitemizethanks{\IEEEcompsocthanksitem Ke Zhang and Xiahai Zhuang are with the School of Data Science, Fudan University, Shanghai, China.
			\IEEEcompsocthanksitem Xiahai Zhuang$^*$ is the corresponding author (zxh@fudan.edu.cn)
			\IEEEcompsocthanksitem The code will be released via https://zmiclab.github.io/projects.html once this manuscript is accepted for publication.
		}
	}
	
	\markboth{IEEE TRANSACTIONS ON PATTERN ANALYSIS AND MACHINE INTELLIGENCE, VOL. XX, NO. XX, 2021}%
	{Shell \MakeLowercase{\textit{et al.}}: Bare Demo of IEEEtran.cls for Computer Society Journals}
	\IEEEtitleabstractindextext{
		\begin{abstract}
Partially-supervised learning can be challenging for segmentation due to the lack of supervision for unlabeled structures, and the methods directly applying fully-supervised learning could lead to \textit{incompatibility}, meaning ground truth is not in the solution set of the optimization problem given the loss function.
To address the challenge, we propose a deep compatible learning (DCL) framework, which trains a single multi-label segmentation network using images with only partial structures annotated. 
We first formulate the partially-supervised segmentation as an optimization problem compatible with missing labels, and prove its compatibility. 
Then, we equip the model with a conditional segmentation strategy, to propagate labels from multiple partially-annotated images to the target. 
Additionally, we propose a dual learning strategy, which learns two opposite mappings of label propagation simultaneously, to provide substantial supervision for unlabeled structures.
The two strategies are formulated into compatible forms, termed as conditional compatibility and dual compatibility, respectively.
We show this framework is generally applicable for conventional loss functions. 
The approach attains significant performance improvement over existing methods, especially in the situation where only a small training dataset is available. 
Results on three segmentation tasks 
have shown that the proposed framework could achieve performance matching fully-supervised models.
		\end{abstract}
		}
	\maketitle
	
\begin{figure}[thb]
		\centering
		\includegraphics[width=0.95\linewidth]{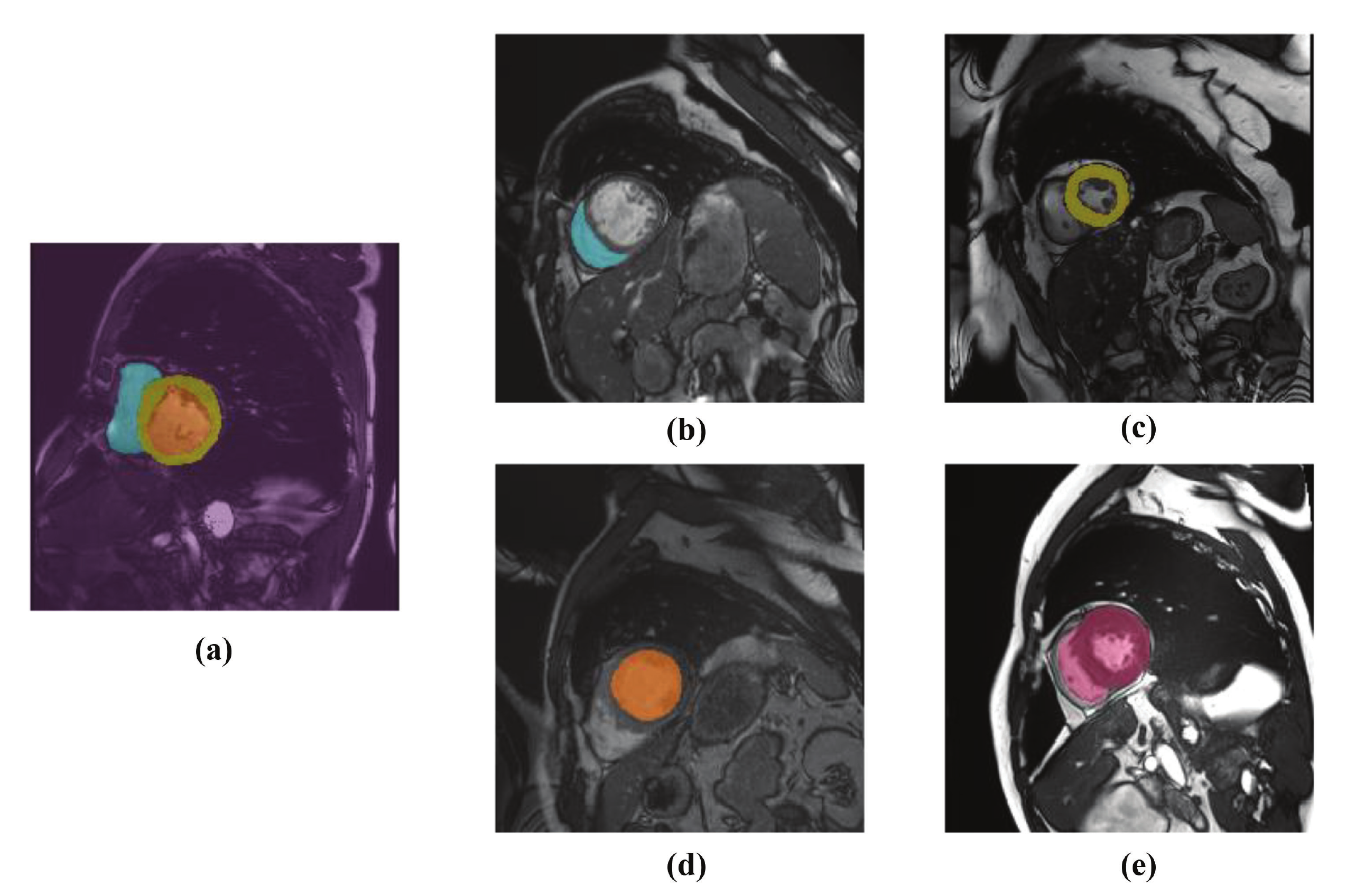}
		\caption{
			Illustration of a fully annotated cardiac MR image (a) and four typical partially annotated images (b)-(e). The colored areas indicate the annotated cardiac structure (blue for right ventricle, orange for left ventricle, yellow for myocardium, pink for whole heart and purple for background). Note that the pink label for  whole heart is equivalent to the partial label of the background (non-heart structure).} 
		\label{fig1}
	\end{figure}

	\IEEEdisplaynontitleabstractindextext
	\IEEEpeerreviewmaketitle
	\IEEEraisesectionheading{\section{Introduction}\label{sec:introduction}}
	\noindent 
	Strong supervision from fully labeled datasets is generally necessary for the success of deep learning based segmentation~\cite{zhao2019data,chen2018voxresnet,yang2017automatic}. 
	However, it can be both costly and expertise-demanding to curate a large-scale fully annotated dataset. Meanwhile, in real-world applications, a great number of datasets are partially annotated to meet different practical usages or research goals.
	This is commonly seen in medical image analysis\cite{wang2019benchmark}.
	For example, for myocardial viability studies, only the myocardial area of the left ventricle (LV) needs to be annotated; while for diseases with abnormal right ventricles (RV), anatomical information of the RV is of interest and would be extracted.
	\zkreffig{fig1} illustrates a fully labeled cardiac image and four partially annotated ones.
	DNN-based methods, which have demonstrated great potential\cite{ronneberger2015u,cciccek20163d,isensee2021nnu}, are mainly designed for supervised learning with training images of full annotation. 
	Hence, the partially annotated images could not be utilized directly or \emph{could be misused}. 
	To make full use of these data, we propose to investigate the scenario of partially-supervised segmentation.
	
	\begin{figure*}[t]
		\centering
		\subfigure[]{
			\includegraphics[width=0.3\textwidth]{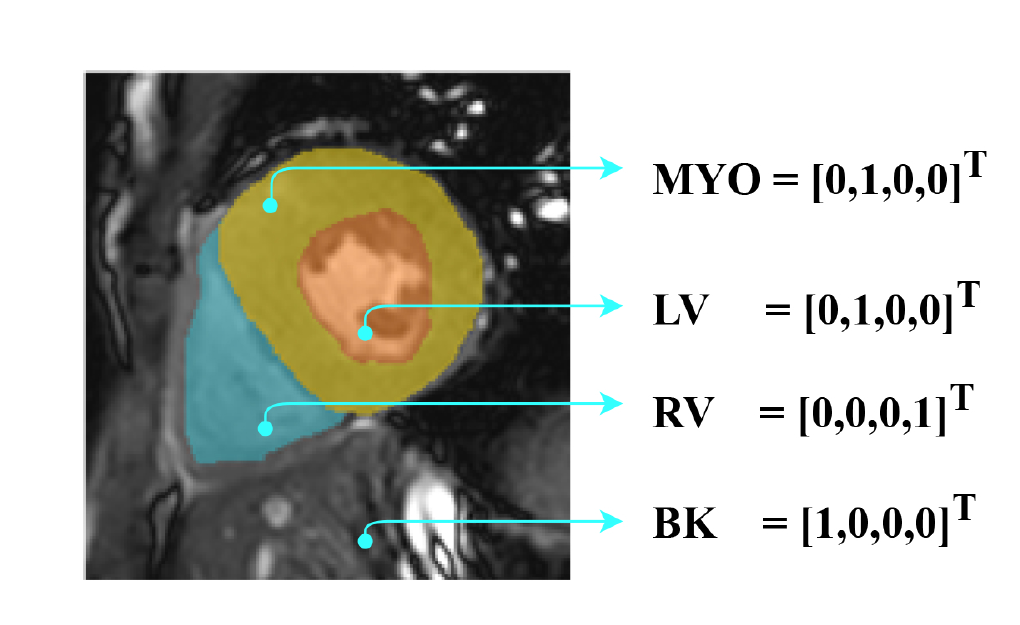}}
		\subfigure[]{
			\includegraphics[width=0.3\textwidth]{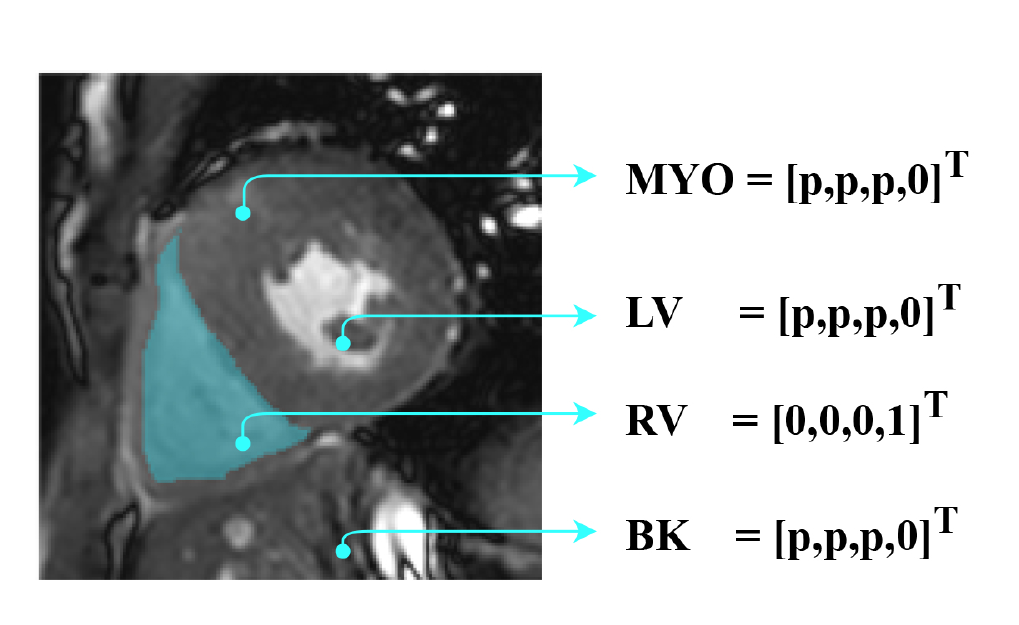}}
		\subfigure[]{
			\includegraphics[width=0.3\textwidth]{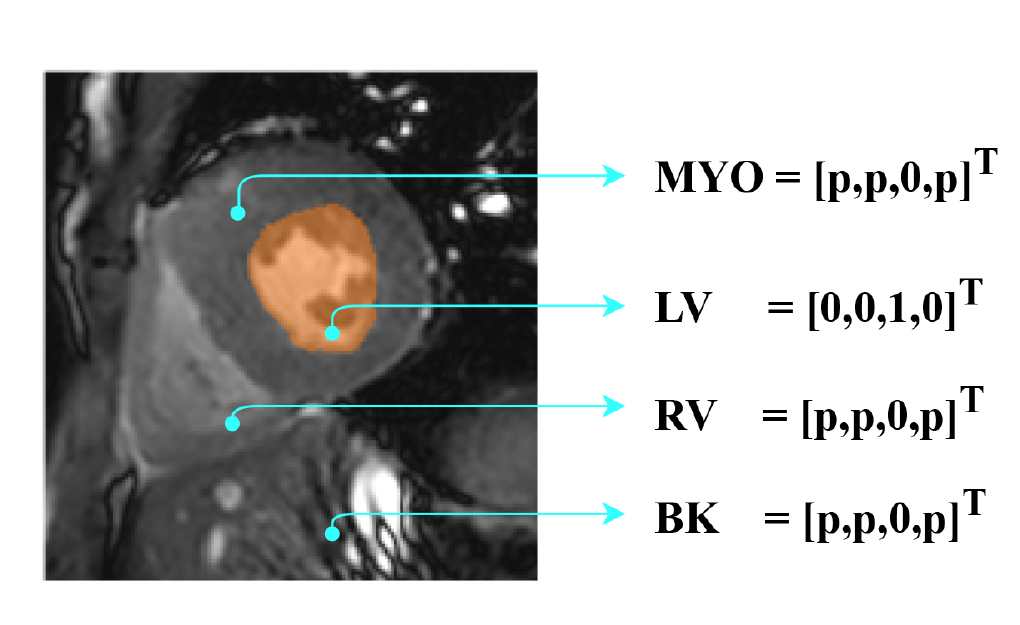}}
		\caption{An image from a fully annotated dataset (a) and its partially annotated cases (b,c). For fully annotated image (a), each pixel has its corresponding label. By contrast, for partially annotated cases, the unlabeled pixels could be represented as different vectors.
			Fig. (b) and (c) illustrate the two forms of partial labels for the unlabeled structures in the partially annotated images. 
			The acronyms are as follows: RV for right ventricle, MYO for myocardium, LV for left ventricle and BK for background.}
		\label{fig2}
	\end{figure*}
	
	In supervised segmentation, there is a ground truth label for each pixel, which should belong to the optimal solution set of a loss function. 
	However, in the case of partially-supervised segmentation, training images are annotated with partial labels, where unlabeled pixels are represented as label vectors indicating their membership of label classes~\cite{wu2015ml}, and the ground truth label may not be in the solution set. 
	Concretely, for the label vector of an annotated pixel, we assign an element of the vector with value of $1$ indicating the pixel being a member of the label class, or with value of $0$ denoting not being a member of the class; 
	while for un-annotated pixels, we assign elements with value of $p$ $(0<p<1)$ meaning that the membership is unknown, or with value of $0$ indicating non-membership.
	Therefore, unlabeled pixels with the same label class or structure can be represented by different label vectors in different images.
	\zkreffig{fig2} provides the illustration, where unlabeled pixels in myocardium (the same label class) are represented by three different values, i.e., $[0,1,0,0]^\mathrm{T}$, $[p,p,p,0]^\mathrm{T}$ and $[p,p,0,p]^\mathrm{T}$.
	This difference of label values can lead to \textit{incompatibility} for conventional pixel-wise segmentation loss functions (such as cross entropy and Dice loss). 
	
	\emph{Incompatibility} means the ground truth label is not in the solution set of the loss function.
	To alleviate it, a straightforward strategy is to directly apply the supervision on the labeled structures, ignoring the unlabeled structures in the training images \cite{dmitriev2019learning,zhou2019prior}. 
	Alternatively, one can consider the missing labels as background, such as in the target adaptive loss \cite{fang2020multi}. 
	However, without a large amount of training data, the lack of supervision on unlabeled structures could easily lead to the problem of over-fitting.  
	
	In the compatibility theory of a linear programming\cite{greenberg2000simultaneous,greenberg1986analysis}, the set of compatible directions is composed of directions for which the primal and dual feasibility conditions hold. 
	\textit{Here, we extend the definition of compatibility to the specific segmentation problem, and define the loss function to be compatible if the ground truth belongs to the set of optima.} 
	We will elaborate on this in detail in the methodology section. 
	In the literature, there is a research topic referred to as as visual compatibility, which has been proposed to deal with fashion recommendation and retrieval. 
	This research is aimed to project two compatible fashion items close to each other in the style space \cite{han2019finet, vasileva2018learning}, for example using the compatibility scoring \cite{han2017learning}. 
	Different from visual compatibility, our topic is related to an optimization problem, for partially-supervised image segmentation with missing labels.
	
	In this work, we focus on compatibility for partial supervised learning to fully exploit supervision on unlabeled structures of the training images. In the case of limited training data with missing labels, it is particularly challenging for an algorithm to find the optimal solution set.
	Therefore, we propose the deep compatible learning (DCL) framework, which jointly incorporates (1) compatibility, (2) conditional compatibility and (3) dual compatibility.
	\emph{For compatibility}, we formulate the partially-supervised segmentation as an optimization problem compatible with missing labels; 
	and we propose a new formulation to derive compatible loss, which is aimed to exploit maximal supervision from partially-annotated training images.  
	\emph{For conditional compatibility}, we leverage the conditional segmentation strategy \cite{hu2019conditional} and formulate it into the compatible form. Considering that target objects in different images usually have an internal connection among context, shape and location~\cite{wang2019pairwise}, we introduce the conditional compatibility to propagate missing labels from other images annotated with these labels. 
	\emph{For dual compatibility}, we incorporate the dual learning strategy and propose its compatible form, termed as dual compatibility, by requiring the label propagation process to be compatible in a closed-loop manner. 
	The dual compatible network is presented to perform the closed-loop scheme by learning two opposite label propagation between the conditional image and the target image simultaneously.
	\emph{Finally}, the proposed framework is generally applicable to any existing loss function originally designed for fully-supervised or partially-supervised segmentation, to boost the performance of these loss functions for partially-supervised segmentation.

	The contributions of this paper are summarized as follows:
	\begin{itemize}
		\item We propose a framework for multi-label segmentation with only partially annotated training images,
		where the learning task is formulated as an optimization problem with a compatible loss. This framework is proved to be \textit{compatible} with missing labels.
		\item We propose to jointly incorporate conditional segmentation and dual learning strategy to endow the algorithm with substantial supervision for missing labels. 
		We further formulate the training strategies into compatible forms, referred to as conditional compatibility and dual compatibility, respectively.
		\item The proposed deep compatible framework could be applied to any existing loss function that is compatible with fully-supervised segmentation. For illustration, we show this framework to be generally applicable for the existing loss functions, including cross entropy, Dice loss, target adaptive loss\cite{fang2020multi} and exclusive loss\cite{shi2020marginal}.
		\item  The proposed deep compatible learning (DCL) has been validated using three segmentation tasks. DCL matches the performance of fully-supervised methods on all three tasks. We use three parameter studies to illustrate the compatibility, data insensitivity, and applicability of DCL. 
	\end{itemize}
	
	This paper is organized as follows: Section \ref{sec02} provides a brief survey to related work, including the topics of partially-supervised segmentation, conditional segmentation and dual learning. In Section \ref{sec03}, we define compatibility and further introduce the conditional compatibility and dual compatibility. Additionally, we show that DCL is generally applicable to existing loss functions. Section \ref{section5} presents results of validation studies. Finally, we conclude in Section \ref{section6}.
	
	\section{Related work}\label{sec02}
	Research literature on incomplete supervision exhibits great diversity, spanning from semi-supervised learning \cite{mittal2019semi, souly2017semi,tarvainen2017mean} to weakly-supervised learning\cite{wei2016stc,pathak2015constrained,khoreva2017simple}. 
	In this work, we focus on the partially-supervised segmentation methods most relevant to our research. In addition, we borrow the idea of conditional segmentation when formulating conditional compatibility. 
	Finally, our dual compatibility is related to the dual learning.
	
	\subsection{Partially-supervised segmentation}
	Existing research works on partially-supervised segmentation are limited, to the best of our knowledge. 
	Dmitriev \textit{et al.}~\cite{dmitriev2019learning} proposed a unified highly efficient framework by utilizing a novel way of conditioning a convolutional network for the purpose of segmentation.
	Zhou \textit{et al.}~\cite{zhou2019prior} incorporated prior knowledge about organ size distribution through a prior-aware loss, which assumes that the average organ size distributions in the abdomen should approximate their empirical distributions.
	However, the methods in \cite{dmitriev2019learning,zhou2019prior} both ignore the cross entropy of unlabeled structures during the back-propagation.
	Fang \textit{et al.}\cite{fang2020multi} developed a pyramid-input and pyramid-output feature abstraction network, and combined it with the target adaptive loss, which treats the unknown labels as background to allow computing the loss.
	Shi \textit{et al.}\cite{shi2020marginal} formulated the target adaptive loss into two forms based on cross entropy and Dice loss.
	They also proposed exclusive loss, to leverage the mutual exclusiveness between different organs. 
	However, the method of taking unlabeled structures as background ignores the category-specific anatomy information.  It is difficult to provide substantive supervision for unlabeled structures, especially in the case of small training data. We highlight that DCL is compatible with the existing state-of-the-art model architectures and loss functions. Furthermore, DCL provides additional substantive supervision for unlabeled structures, from the perspective of compatibility.
	
	There is a research topic referred to as partially-supervised instance segmentation~\cite{hu2018learning,kuo2019shapemask,fan2020commonality}, which aims to train instance segmentation models on limited mask-annotated categories of data, and generalize the models to new categories with only bounding-box annotations available. Different from this partially-supervised instance segmentation, our topic focuses on training a multi-label segmentation network using images with only partial structures annotated.
	
	
	
	\subsection{Conditional segmentation}
	Considering that the labels in different images are spatially and physically interrelated, the conditional images, also referred to as moving images or atlases, are taken as input to provide prior knowledge, which is termed as conditional priors~\cite{okada2015abdominal}, for neural networks.
	In recent literature, conditional segmentation~\cite{hu2019conditional} propagated labels on a moving image to a fixed image directly.
	Other recent methods, such as pairwise segmentation~\cite{wang2019pairwise,wang2020pairwise} and co-segmentation~\cite{dorent2020scribble},  utilized a pair of samples to facilitate the prediction of pixel-level masks.
	Dorent \textit{et al.}~\cite{dorent2020scribble} formulated the domain adaptation as a co-segmentation task and presented a structured learning approach to propagate information across domains. 
	Wang \textit{et al.}~\cite{wang2019pairwise,wang2020pairwise} proposed a proxy supervision in pairwise image segmentation to address intra-class heterogeneity and boundary ambiguity across a pair of samples. 
	However, it relies on a fusion net to exploit two streams of input.
	Different from these methods that segment multiple images simultaneously \cite{dorent2020scribble,wang2019pairwise,wang2020pairwise}, the proposed conditional compatibility is based on conditional segmentation~\cite{hu2019conditional}, and focuses on predicting the segmentation results of the target image by propagating labels from multiple conditional images to the target image.
	Furthermore, it is worth mentioning that none of the above works considers the case of partial supervision. 
	To deal with missing labels, we propose a novel loss term for priors, which is designed to exploit the explicit inclusiveness and exclusiveness relationships between the outputs and the labels of conditional images.

	\subsection{Dual learning}
	Dual learning methods~\cite{he2016dual,xia2017dual,xia2018model,zhang2018deep} exploit the probabilistic correlation between two dual tasks to regularize the training process. Generally, dual learning~\cite{xia2017dual} involves a primal task and a dual task. For the two tasks, the input and output of one task are exactly the output and input of the other. Dual learning was originally applied to supervised learning tasks in dual forms such as translation between two languages, and was recently used to leverage such dualities in image translation~\cite{zhu2017unpaired,yi2017dualgan} and image super-resolution~\cite{guo2020closed}. These tasks are easily represented as dual forms, such as target domain$\leftrightarrow$source domain, low-resolution images$\leftrightarrow$high-resolution images. However, there is no inherent duality in the image segmentation task. Since the segmentation mask does not contain all the information of the image, it is infeasible to learn the pixel-mapping from labels to images. To solve this problem, we design the dual tasks based on the assumption that networks can learn two opposite mappings of label propagation simultaneously, from the conditional image to the target image, and vise versa.
	
	\section{Method}\label{sec03}
	
	\begin{table}[t]
		\centering 
		\caption{Reference to the mathematical symbols} 
		\label{tab1}
		\begin{tabular}{|c|p{6cm} |}
			\bottomrule
			Notations & Notions\\
			\hline
			$\mathcal{S}_P$ & partially annotated dataset\\ 
			$\bm{x},\bm{y}$ & image, segmentation of image\\ 
			$h,w,d$ & image height, width and depth\\ 
			$\Omega; V$ & set of pixel in an image; number of pixels\\
			$i,x_i$& index of a pixel, the pixel with index $i$\\
			$\bm{y}_i,\hat{\bm{y}}_i,\bm{e}_i$&label vector, prediction, ground truth label of $x_i$\\
			\hline 
			$m$ & number of label classes\\				
			$\bm{c}\!=\!\!\{c_j\}^m_{j=1}$ & set of label classes or label class indices\\
			$c(x_i),\bar{\bm{c}}_i$ & label class index of $x_i$, the complementary set of label classes ($\bm{c}-\{c(x_i)\}$)\\
			$\bm{c}_q,\bar{\bm{c}}_q$ &set of known label classes in a partially labeled image, set of unknown label classes\\
			$\Omega_q,\bar{\Omega}_q$& set of labeled pixel indices, set of unlabeled pixel indices\\
			\hline
			Superscript$^{t/c}$& variables of target image/ conditional image\\
			Superscript$^{\cap/\Delta}$& intersection/ extra part\\
			$g(\cdot)$ & conditional supervision\\
			$\bm{z}_i$ & label vector from conditional supervision of pixel $x_i$\\
			\hline
			Subscript$_{p/d}$ & variables of primal/ dual network\\
			$P(\cdot),D(\cdot)$& primal mapping, dual mapping\\
			$s$& randomly selected label class for dual compatibility\\
			\toprule
		\end{tabular}\label{notation}   
	\end{table}
	
	This section introduces the DCL framework for partially-supervised multi-label segmentation. 
	The important notations and descriptions are summarized in Table~\ref{tab1}. 
	Firstly, we give a formal definition of compatibility and propose a formulation for compatible loss to exploit supervision from partial labels, in Section~\ref{method:compatibility}.
	Then, in Section~\ref{section3.2} we introduce the conditional compatibility, which is aimed to propagate labels from multiple partially annotated images to the target image. This is achieved via a DNN, which is trained with a compatible loss and is referred to as CompNet. 
	Section~\ref{section3.3} describes the dual compatibility, which is aimed to  form the closed-loop label propagation. 
	The  dual compatibility is achieved using a compatible network composed of the primal network (PrimNet) and the dual network (DualNet). 
	Finally, we show applications of DCL to existing loss functions in Section \ref{section3.4}. 
	
	\begin{figure*}[tb] \centering 
		\includegraphics[width=0.8\textwidth]{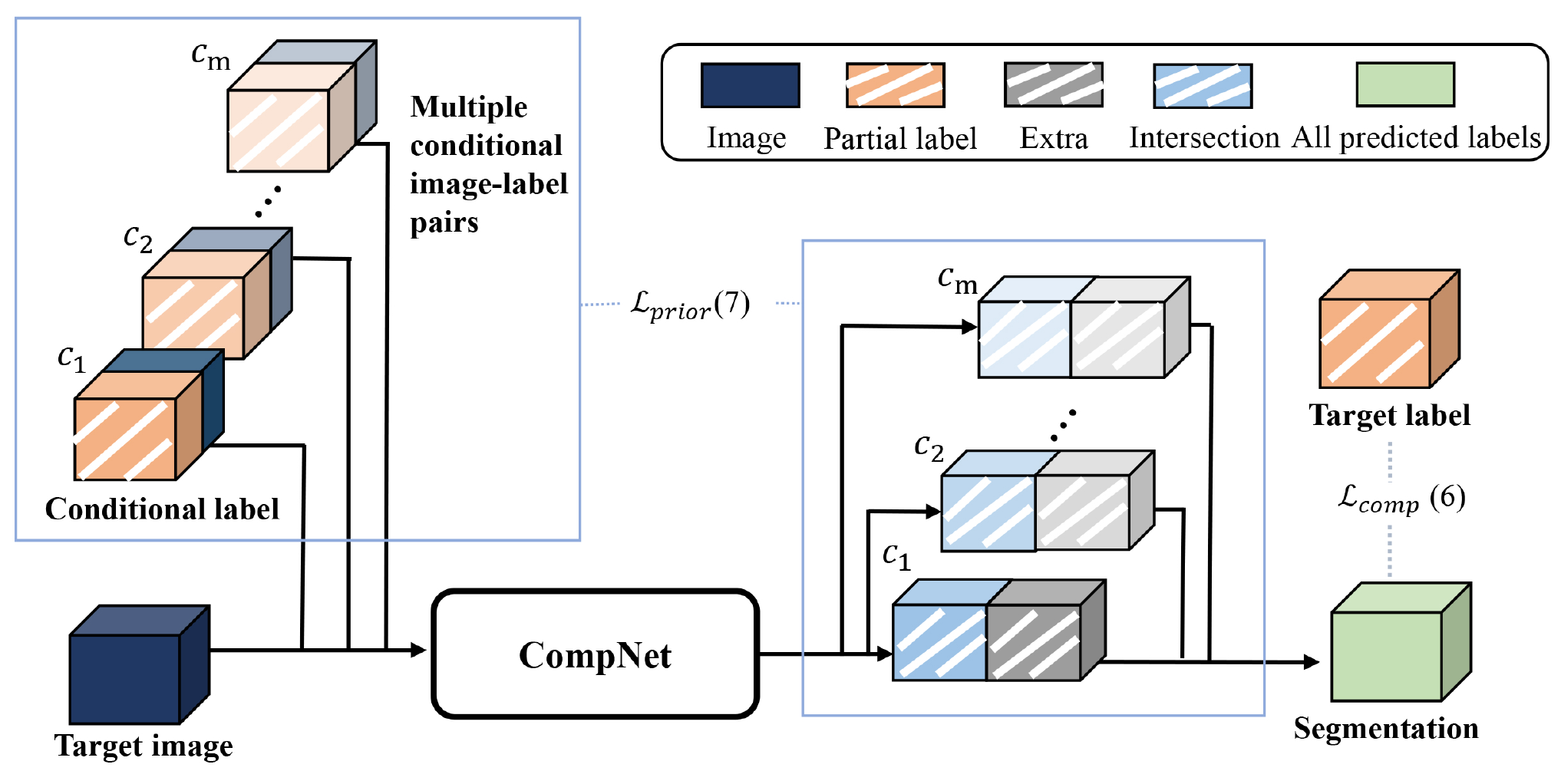}
		\caption{Overview of the compatible network (CompNet) using conditional compatibility. CompNet propagates labels from multiple partially annotated conditional images to the target image.
		}  
		\label{fig3}
	\end{figure*}

	\subsection{Compatibility}\label{method:compatibility}
	
	Let $\bm{x}$ be an image of size $h\times w \times d$, namely with $V$ pixels $\{x_i|_{i \in \{1,\cdots, V\}}\}$ and $V = h\times w \times d$.
	The segmentation has $m$ label classes, \textit{i.e.}, $\{c_1,\cdots, c_m\}$, and $\bm{y}_i$ is the label value of pixel ${x}_i$.
	A training image may only have $q$ classes labeled and the set of labeled class indices is denoted as $\bm{c}_q$, leaving other $(m-q)$ classes unlabeled, denoted as $\bar{\bm{c}}_q$. Note that when $q=m-1$, it is equivalent to the fully annotated case. Therefore, partial annotation requires $m-q\geq 2$ and $m\geq 3$. 
	Let $\Omega_q$ be the set of annotated pixel indices in $\bm{x}$, and $\bar{\Omega}_q$ be the set of remaining unlabeled pixel indices.
	For pixel $x_i$ whose ground truth label class is $c_j$, \textit{i.e.}, $c(x_i)=c_j$. If $x_i$ is annotated ($i\in \Omega_q$), its label value is denoted using a one-hot vector $\bm{y}_i = [y_{i1},y_{i2},\cdots, y_{im}]$ with $y_{ij} =1$ and the others being 0;
	otherwise ($i\in \bar{\Omega}_q$), its label value can be represented as $\bm{y_i} \in \{0,p\}^{1 \times m}$, where $y_{ij} = 0$ means $x_i$ is not a member of $c_j$, and $y_{ij} = p$ indicates the membership of $x_i$ to $c_j$ is missing, and $p\in (0,1)$ is a probability value.
	In the following, we first define the concept of compatibility and then propose a formulation for deriving a compatible loss.  
	
	\subsubsection{Definition of Compatibility}
	Definition~\ref{definition1} formulates \textit{compatibility} for partially-supervised learning. 
	If a loss function $\mathcal{L}$ is incompatible, the ground truth label is not in the solution set obtained via minimization of $\mathcal{L}$ using partial labels from partially annotated training set.
	For fully-supervised segmentation tasks, the conventional loss functions (such as cross-entropy) are generally compatible, indicating the ground truth of all pixels belongs to the solution set of the loss function, as Definition~\ref{definition1} requires.  
	
	\begin{definition}[Compatibility of loss functions] 
		Let $\bm{x}$ be the image with $V$ pixels. Each pixel of $\bm{x}$ is partially annotated with a vector of dimension $1\times m$. The ground truth of each pixel is a one-hot vector. Denote the partial label of image as $\bm{a}$ and the ground truth label as $\bm{e}$, which are of size $V \times m$. For minimization of loss function $\mathcal{L}$, the solution space $\mathcal{D}_a$ for $\bm{a}$ is formulated as:
		\begin{equation}
			\begin{aligned}
				\mathcal{D}_a = \mathop{\arg\min}\limits_{\hat{\bm{a}}}\mathcal{L}(\hat{\bm{a}}, \bm{a})\\
			\end{aligned}.
		\end{equation}
		If $\bm{e} \in \mathcal{D}_a$, the loss function $\mathcal{L}$ and the optimization problem are defined to be compatible over $\bm{a}$.
		\label{definition1}
	\end{definition}
	
	In partially-supervised learning, unlabeled pixels of the same class can be represented by different values of label vectors in different images, as we show in \zkreffig{fig2}. 
	This could lead to incompatibility of the loss. In the following, we prove cross entropy loss (CE) is not compatible in this case.
	
	\begin{proposition}[Incompatibility of conventional CE loss]
		Image $\bm{x}$ is partially annotated with $\bm{c}_{q}$, where $1\leq q \leq m-2$. The partial label of $\bm{x}$ is represented as $\bm{a} \in \{0,p,1\}^{V\times m}$.
		For pixel-level classification, the probability of each class is bounded in $[0,1]$, and the sum of the probabilities over all classes equals 1. Then, the optimization problem is incompatible using cross entropy loss ($\mathcal{L}_{ce}$) for minimization.
		\begin{proof}
			The ground truth label of $\bm{x}$ is denoted as $\bm{e}$. 
			For each pixel $x_i$, the label $\bm{e}_i$ is a one-hot vector.
			If $j = c(x_i)$, $\bm{e}_{ij} = 1$.
			Otherwise, $e_{ij} = 0$.
			If $i \in \Omega_q$, $\bm{a}_{i}=\bm{e}_{i}$.
			If $i \in \bar{\Omega}_q$, the elements of partial label vector $\bm{a}_i$, denoted as $\{a_{ij}|_{j=1,\cdots,m}\}$, we have $a_{ij}=0$ for $j\in \bm{c}_q$ and $a_{ij}=p$ for $j\in \bar{\bm{c}}_q$.
			Let $\mathcal{D}_a$ be the solution set of $\hat{\bm{a}}$, \textit{i.e.}, $\mathcal{D}_{a} =\mathop{\arg\min}\limits_{\hat{\bm{a}}}\mathcal{L}_{ce}(\hat{\bm{a}},\bm{a})$.
			Since minimizing $\mathcal{L}_{ce}$ is a convex optimization problem~\cite{chen2014convolutional}, it has an optimal solution set $\mathcal{D}_a = \{\bm{a}\}$, which means that $\bm{e} \notin \mathcal{D}_a$.
			Therefore, conventional CE loss and its optimization problem are incompatible in partially-supervised segmentation.
		\end{proof}
		\label{proposition1}
	\end{proposition} 
	Similarly, one can prove that the Dice loss, which is widely used for fully-supervised learning, is incompatible in partially-supervised segmentation. 
	
	\subsubsection{A formulation for compatible loss}
	
	We propose a new formulation to derive a compatible loss, as follow,
	\begin{equation}
		\mathcal{L}_{comp}(\hat{\bm{y}},\bm{y}) =  \alpha_1\mathcal{L}_P(\hat{\bm{y}},\bm{y}) + \alpha_2\mathcal{L}_N(\hat{\bm{y}},\bm{y}) ,
		\label{eq2}
	\end{equation} 
	which is composed of two loss functions, \textit{i.e.}, the positive  loss ($\mathcal{L}_P$) and the negative loss ($\mathcal{L}_N$), 
	and $\alpha_1 > 0$ and $\alpha_2 >0$ are balancing parameters.  
	These two functions are designed to satisfy certain conditions, as follows, 
	\begin{equation}
		\begin{aligned}
			\mathcal{L}_P(\hat{\bm{y}},\bm{y}) = \sum_{i=1}^{V}\sum_{j=1}^m \mathbbm{1}_{[y_{ij}=1]} f_P(\hat{y}_{ij}, y_{ij}) , \\
			\text{s.t.\ } f_P(1,1) \leq f_P(a, 1),\ \forall a \in [0,1] ;
		\end{aligned} 
		\label{eq-P}
	\end{equation}
	\noindent and,
	\begin{equation}
		\begin{aligned}
			\mathcal{L}_N(\hat{\bm{y}},\bm{y}) = \sum_{i=1}^{V}\sum_{j=1}^m \mathbbm{1}_{[y_{ij}=0]} f_N(\hat{y}_{ij}, y_{ij}) , \\
			\text{s.t.\ } f_N(0,0) \leq f_N(a, 0),\ \forall a \in [0,1] ,
		\end{aligned} 
		\label{eq-N}
	\end{equation} 
	where $f_P(\hat{y}_{ij}, y_{ij})$ and $f_N(\hat{y}_{ij}, y_{ij})$ measure the non-negative distance between $\hat{y}_{ij}$ and $y_{ij}$ for pixel $x_i$. 
	If the label information regarding $c_j$ is certain, \textit{i.e.}, either $y_{ij}=1$ or $y_{ij}=0$, both of the two distances, \textit{i.e.}, $f_P$ and $f_N$ can be minimized by predicting $\hat{y}_{ij}= y_{ij}$.
	
	In the following, we first prove both of the positive and negative loss functions are compatible; then, we prove compatibility satisfies linearity, based on which we prove that the proposed formulation in Eq.~(\ref{eq2})  is a compatible loss.
	
	\begin{proposition}[Compatibility of $\mathcal{L}_P$ and $\mathcal{L}_N$] 
		The two loss functions defined in Eq.~(\ref{eq-P}) and (\ref{eq-N}) are compatible losses. 
		\begin{proof} 
			For pixel $x_i$ in image $\bm{x}$, $\bm{y}_i$, $\bm{\hat{y}}_i$ and $\bm{e}_i$ are respectively its label vector, prediction and ground truth. 
			Note that only the pixels whose label indices belong to the annotated label set $\bm{c}_q$ can have an element value being 1 in their label vectors.  
			Therefore, $\mathcal{L}_P$ can be rewritten as,
			$\mathcal{L}_P(\hat{\bm{y}},\bm{y}) = \sum_{i\in \Omega_q}\sum_{j\in\bm{c}_q} f_P(\hat{y}_{ij}, 1)$. 
			From Eq.~(\ref{eq-P}), we have 
			\begin{displaymath}\begin{aligned}
					\mathcal{L}_P(\bm{e},\bm{y})&=\sum_{i\in \Omega_q}\sum_{j\in\bm{c}_q} f_P(1, 1) \\ 
					&\leq \sum_{i\in \Omega_q}\sum_{j\in\bm{c}_q} f_P(\hat{y}_{ij}, 1)
					=\mathcal{L}_P(\hat{\bm{y}},\bm{y}),\ \forall \hat{y}_{ij}\in[0,1]  .
			\end{aligned}\end{displaymath}
			Therefore, we have  $\mathcal{L}_P(\bm{e},\bm{y})$$\leq$$\mathcal{L}_P(\hat{\bm{y}},\bm{y})$, $\forall \hat{\bm{y}}$ whose $\hat{y}_{ij}\in[0,1]$.
			Let $\mathcal{D}_P = \arg\min_{\hat{\bm{y}}}\mathcal{L}_P(\hat{\bm{y}},\bm{y})$, we have $\bm{e}\in\mathcal{D}_P$. 
			Therefore, $\mathcal{L}_P$ is compatible according to Definition 1.
			
			Similarly, $\mathcal{L}_N$ can be rewritten as,
			\begin{displaymath}
				\mathcal{L}_N(\hat{\bm{y}},\bm{y}) = 
				\sum_{i\in \Omega_q}\sum_{j\in \bar{\bm{c}}_i} f_N(\hat{y}_{ij}, 0)+
				\sum_{i\in\bar{\Omega}_q}\sum_{j\in \bm{c}_q} f_N(\hat{y}_{ij},0)  , 
			\end{displaymath}
			where, $\bar{\bm{c}}_i$ denotes the set of class indices that excludes the class of $x_i$.
			From Eq.~(\ref{eq-N}), we have 
			\begin{displaymath}\begin{aligned}
					\mathcal{L}_N(\bm{e},\bm{y})&= 
					\sum_{i\in \Omega_q}\sum_{j\in \bar{\bm{c}}_i} f_N(0, 0)+
					\sum_{i\in\bar{\Omega}_q}\sum_{j\in \bm{c}_q} f_N(0,0) \\
					&\leq \mathcal{L}_N(\hat{\bm{y}},\bm{y}),\  \forall \hat{\bm{y}} \text{ whose } \hat{y}_{ij}\in[0,1].
			\end{aligned}\end{displaymath} 
			Let $\mathcal{D}_N= \arg\min_{\hat{\bm{y}}}\mathcal{L}_N(\hat{\bm{y}},\bm{y})$, we have $\bm{e}\in\mathcal{D}_N$. Therefore, $\mathcal{L}_N$ is compatible according to Definition 1.
		\end{proof}
		\label{propositionpixelpn}
	\end{proposition}
	
	\begin{proposition}[Linearity of compatibility] 
		Assume that $\mathcal{L}_{1}$ and $\mathcal{L}_{2}$ both be compatible loss functions, their weighted sum, $\mathcal{L}_w = w_1 \mathcal{L}_{1}+w_2\mathcal{L}_{2}$, where $w_1>0$ and $w_2>0$, is  compatible. 
		\begin{proof}
			Denote the ground truth label of image $\bm{x}$ as $\bm{e}$. 
			For any partial labeling form $\bm{a}$, let the solution spaces of $\mathcal{L}_{1}, \mathcal{L}_{2}, \mathcal{L}_w$ be $\mathcal{D}_{1}, \mathcal{D}_{2}, \mathcal{D}_{w}$, respectively. 
			According to the definition of compatibility, we have $\bm{e} \in \mathcal{D}_{1}$ and $\bm{e} \in \mathcal{D}_{2}$. 
			Therefore, for any predicted label $\hat{\bm{a}}$, $\mathcal{L}_1(\bm{e}, \bm{a}) \leq \mathcal{L}_1(\hat{\bm{a}}, \bm{a})$ and  $\mathcal{L}_2(\bm{e}, \bm{a}) \leq \mathcal{L}_2(\hat{\bm{a}}, \bm{a})$. 
			Considering that $\mathcal{L}_w (\bm{e}, \bm{a}) =  w_1 \mathcal{L}_{1}(\bm{e}, \bm{a})+w_2\mathcal{L}_{2}(\bm{e}, \bm{a}) \leq  w_1 \mathcal{L}_{1}(\hat{\bm{a}}, \bm{a})+w_2\mathcal{L}_{2}(\hat{\bm{a}}, \bm{a}) = \mathcal{L}_w (\hat{\bm{a}}, \bm{a}) $, we have 
			$\bm{e} \in \mathcal{D}_w$.
			This implies that $\mathcal{L}_w$ is compatible, and the optimization problem of $\mathcal{L}_w$ is compatible.
		\end{proof}
		\label{proposition3}
	\end{proposition}

	Based on Proposition~\ref{propositionpixelpn} and Proposition~\ref{proposition3}, we can prove that the proposed $\mathcal{L}_{comp}$ in Eq.~(\ref{eq2}) is a compatible loss, and  the optimization problem is compatible with missing labels.  
	
	
	\subsection{Conditional compatibility} 	\label{section3.2}

	\zkreffig{fig3} illustrates CompNet for achieving conditional compatibility, which is aimed to propagate labels from multiple partially annotated conditional images to the target image. 
	For simplification, we denote the prior knowledge as conditional priors~\cite{okada2015abdominal}, which are learned by the network in the process of label propagation. 
	The partially annotated images and their labels, used to provide the conditional priors, are denoted as conditional images and conditional labels, respectively.

	\begin{figure}[t]
		\includegraphics[width=\linewidth]{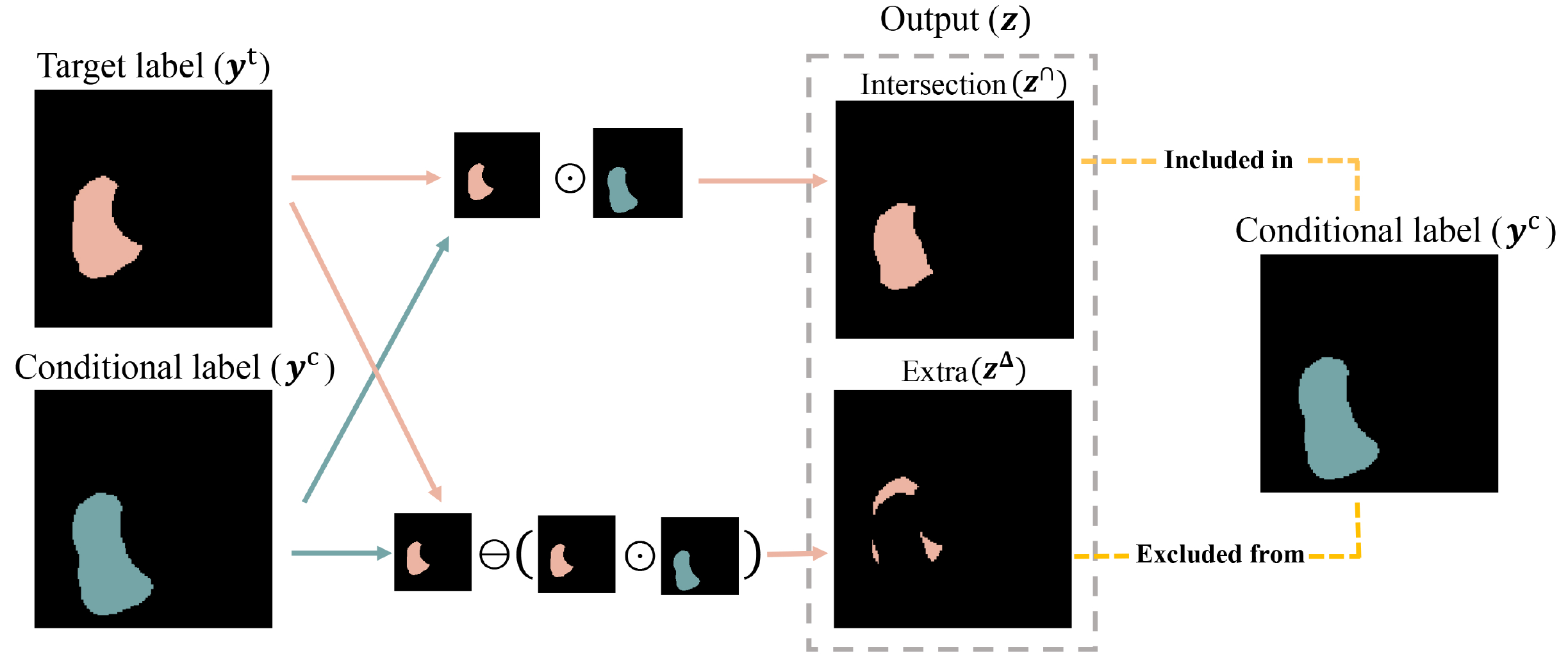}
		\caption{The illustration of the conditional supervision, which exploits the inclusiveness and exclusiveness between output $\bm{z}$ and the conditional label $\bm{y}^c$.  Output $\bm{z}$ consists of intersection and extra. Intersection denotes the intersection between the target label and the conditional label. Extra implies the extra part of the target label minus the intersection. Therefore, intersection should be included in the conditional label, while extra should be excluded from the conditional label. }
		\label{fig5}
	\end{figure}
	
	Let $\bm{x}^t$ and $\bm{y}^t$ denote the target image and its partial label; $\bm{x}^{c_j}$ and $\bm{y}^{c_j}$ indicate a conditional image partially annotated with class $c_j$ and its partial label, respectively.
	CompNet takes $\bm{x}^t$ and $\{(\bm{x}^{c_j}, \bm{y}^{c_j})\}_{j=1}^m$ as input and predicts the multi-label segmentation of the target image as output. 
	To deal with missing labels, we design the conditional supervision to leverage the inclusiveness and exclusiveness between the outputs and conditional labels, as \zkreffig{fig5} shows.
	\textit{Instead of predicting the segmentation directly, CompNet predicts the intersection between the target label and the conditional label, and the extra part of the target label minus the intersection.}
	In other words, we design conditional compatibility with an additional 2-category segmentation task, where each label class consists of two maps, \textit{i.e.}, one for the intersection and the other for the extra part.

	For a pixel at position $i$, the output of CompNet, denoted as $\bm{z}_{i} = [\bm{z}^{\cap}_{i1}, \cdots, \bm{z}^{\cap}_{im},\bm{z}^{\Delta}_{i1}, \cdots, \bm{z}^{\Delta}_{im}]$, is of dimension $1\times2m$, as follows,
	\begin{equation}
		\bm{z} =g(\bm{y}^t,\bm{y}^c) = [\underbrace{\bm{y}^t\odot \bm{y}^c}_{\bm{z}^{\cap}}, \underbrace{\bm{y}^t \ominus(\bm{y}^t\odot \bm{y}^c)}_{\bm{z}^{\Delta}}],
	\end{equation}
	where the label of intersection is represented as $\bm{z}^{\cap}$, and the label of extra part as $\bm{z}^{\Delta}$;
	$\odot$ is the element-wise multiplication operation, and $\ominus$ denotes the element-wise subtraction operation. 
	By simply adding the intersection to the extra part, one can obtain the final segmentation result from the prediction, \textit{i.e.}, $\hat{\bm{y}} = \hat{\bm{z}}^{\cap}\oplus\hat{\bm{z}}^{\Delta}$.
	Therefore, we formulate the compatible loss for $\bm{z}$ and $\hat{\bm{z}}$, as follows,
	\begin{equation}\begin{array}{r@{\ }l}
			\mathcal{L}_{comp}(\hat{\bm{z}},\bm{z})
			&=\mathcal{L}_{comp}(\hat{\bm{z}}^{\cap}\oplus\hat{\bm{z}}^{\Delta},\bm{z}^{\cap}\oplus\bm{z}^{\Delta}) \\
			&=\mathcal{L}_{comp}(\hat{\bm{y}},\bm{y}), \end{array}
		\label{eq4}
	\end{equation} 
	where $\mathcal{L}_{comp}(\hat{\bm{y}},\bm{y})$ is the compatible loss defined in Eq.~(\ref{eq2}), hence $\mathcal{L}_{comp}(\hat{\bm{z}},\bm{z})$ is compatible. 
	
	To leverage conditional priors, we make use of the explicit relationships of inclusiveness and exclusiveness among $\bm{z}^{\cap}$, $\bm{z}^{\Delta}$ and $\bm{y}^c$, as shown in \zkreffig{fig5}. The inclusiveness and exclusiveness mean that the intersection $\bm{z}^{\cap}$ should be included in its conditional label $\bm{y}^c$, while the extra $\bm{z}^{\Delta}$ should be excluded from the conditional label $\bm{y}^c$.  Therefore, the prior loss $\mathcal{L}_{prior}$ is formulated as:
	\begin{gather}
		\small
		\mathcal{L}_{prior}(\hat{\bm{z}}, \bm{y}^c) =\sum_{i=1}^V\sum_{j=1}^m[\underbrace{\mathbbm{1}_{[y^c_{ij}=0]}f_N(\hat{z}^{\cap}_{ij},0)}_{\text{inclusiveness}} +\underbrace{\mathbbm{1}_{[y^c_{ij}=1]}f_N(\hat{z}^{\Delta}_{ij},0)}_{\text{exclusiveness}}],
		\label{eq5}
	\end{gather}
	where the term on the right hand side of above equality consists of two components, \textit{i.e}, the inclusiveness and the exclusiveness. Considering that $\bm{z}^\cap$ needs to be included in $\bm{y}^c$, if $y_{ij}^c$ is 0, the ground truth of $\hat{z}^{\cap}_{ij}$ is also 0. Similarly, since $\bm{z}^{\Delta}$ and $\bm{y}^c$ are mutually exclusive, if $y_{ij}^c$ is 1, the ground truth of $\hat{z}^{\Delta}_{ij}$ is 0. The compatibility of prior loss is proved in Appendix A.
	
	\begin{figure}[t]
		\includegraphics[width=\linewidth]{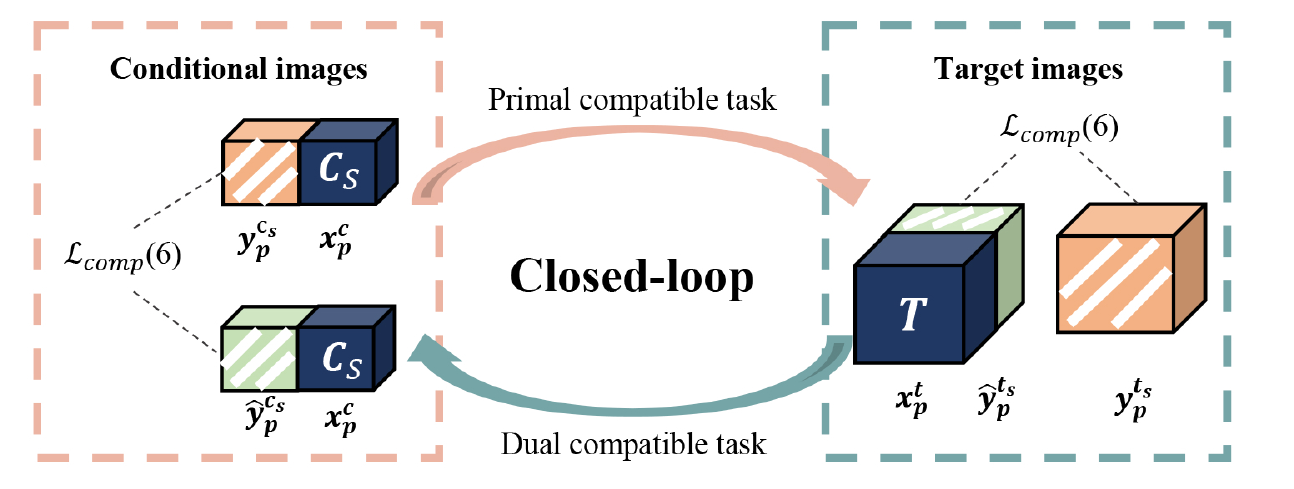}
		\caption{Dual compatibility closed-loop scheme, which contains a primal compatibility task to propagate labels from the conditional images to the target image, and a dual compatibility task to project labels back to the selected conditional image.}
		\label{fig6}
	\end{figure}
	
	Finally, we propose the conditional compatible loss, which is the sum of the compatible loss and the conditional prior loss, as follows,
	\begin{equation}
		\mathcal{L}_{cc}(\hat{\bm{z}},\bm{z}, \bm{y}^c) = \mathcal{L}_{comp}(\hat{\bm{z}}, \bm{z})+\mathcal{L}_{prior}(\hat{\bm{z}}, \bm{y}^c).
		\label{eq6}
	\end{equation}
	Since $\mathcal{L}_{prior}$ and $\mathcal{L}_{comp}$ are both compatible, $\mathcal{L}_{cc}$ is compatible according to Proposition~\ref{proposition3}.
	
	\subsection{Dual compatibility}
	\label{section3.3}
	The dual compatibility requires that the label propagation process needs to be compatible in a closed-loop manner, to enhance the performance of the segmentation network. We first analyze the closed-loop scheme of label propagation. Then, we introduce the dual compatible network, which consists of two CompNets. Finally, we introduce the strategy of network training.
	
	\subsubsection{Closed-loop scheme} 	\label{sec3.3.1}
	
	By swapping the primal target image with the selected conditional image, we form the closed-loop of label propagation on the selected class $c_s$. In \zkreffig{fig6}, we elaborate on the closed-loop setting by introducing two compatibility tasks, \textit{i.e}, the primal compatibility task and the dual compatibility task. 
	We define the primal compatibility task to find the pixel-level mapping $P(\cdot)$ from the selected conditional label $y^{c_s}_{p}$ to the target partial label $y^{t_s}_{p}$, and the dual compatibility task to find a mapping $D(\cdot)$ from $y^{t_s}_{p}$ back to $y^{c_s}_p$. 
	However, considering that the label is determined by its corresponding image, the model needs to learn the pixel-level mapping between labels with  image information.
	Therefore, we modify the primal mapping $P(\cdot)$ and dual mapping $D(\cdot)$ to involve the image information. 
	Let $\bm{\tilde{I}}=\{\bm{x}^{c_s}_p, \bm{x}^t_p\}$, we then formulate the closed-loop scheme as follows,
	\begin{equation}
		\underbrace{(1-\lambda)\mathcal{L} (P(\bm{y}^{c_s}_{p};\bm{\tilde{I}}), \bm{y}^{t_s}_{p})}_{\text{primal compatibility task}} + 
		\underbrace{\lambda\mathcal{L} (D(P(\bm{y}^{c_s}_{p};\bm{\tilde{I}});\bm{\tilde{I}}), \bm{y}^{c_s}_{p})}_{\text{dual compatibility task}},
		\label{eq8}
	\end{equation}
	where $\mathcal{L}(\cdot)$ is a compatible loss to encourage the predicted labels to be consistent with its annotated partial labels;
	$\lambda$ is the balancing parameter. 
	
	We provide a theoretical analysis which shows that the closed-loop scheme has a smaller generalization error bound than the traditional segmentation method. Please refer to Appendix B for proof details. 
	

	\begin{figure*}[t]\centering 
		\includegraphics[width=0.9\textwidth]{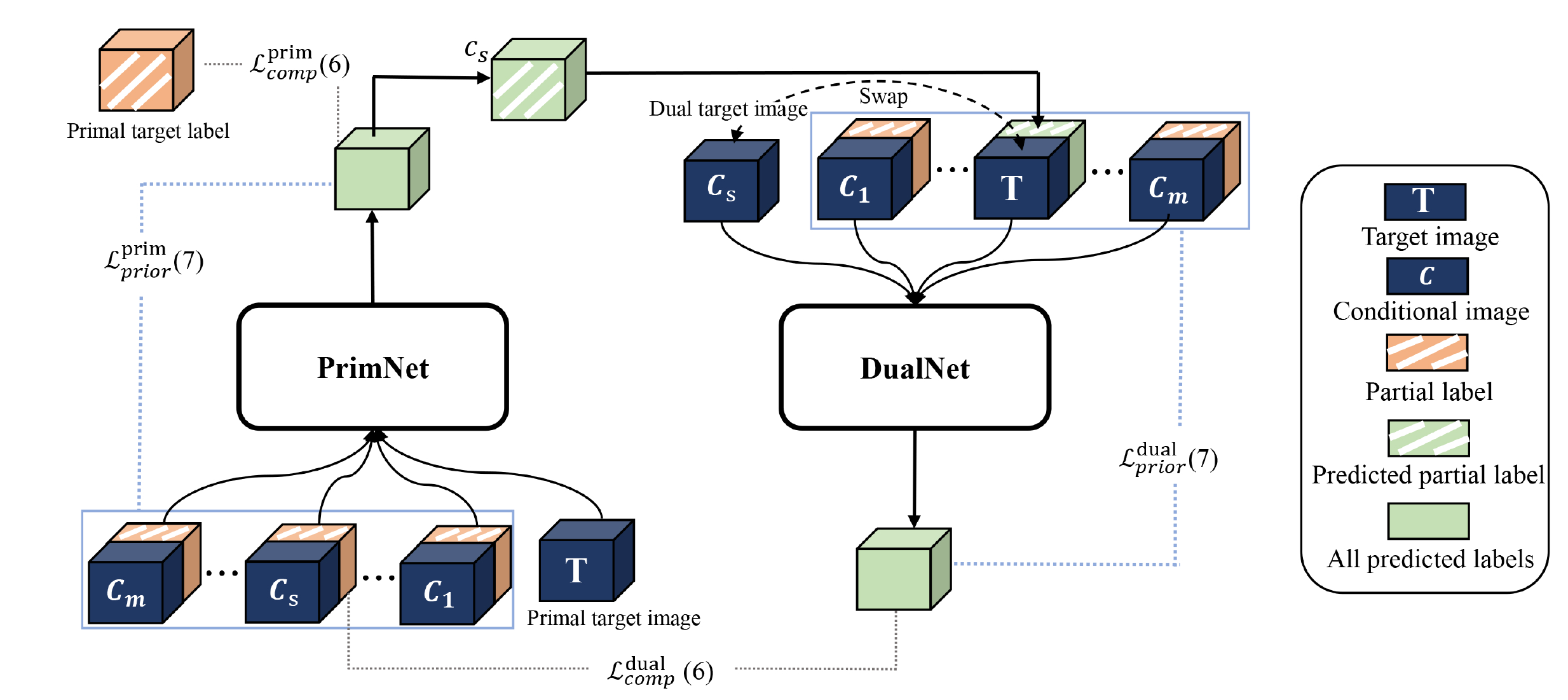}
		\caption{The architecture of the dual compatible network. Dual compatible network contains a primal network (PrimNet) and a dual network (DualNet). Both PrimNet and DualNet adopt the architecture of CompNet. To simplify the figure, the step of outputing the intersection and extra parts is omitted. Therefore, the conditional labels in the blue boxes are directly compared with the network output to calculate the prior losses, \textit{i.e}, $\mathcal{L}^{prim}_{prior}$ of PrimNet and $\mathcal{L}^{dual}_{prior}$ of DualNet, which are represented by the blue dashed lines.}
		\label{fig4} 
	\end{figure*}
	
	\subsubsection{Dual compatible network} 	\label{sec3.3.2}
	We implement the closed-loop scheme with two compatible networks, \textit{i.e.}, PrimNet and DualNet in \zkreffig{fig4}. 
	We denote the variables of PrimNet and DualNet with subscripts of $p$ and $d$, respectively. 
	As \zkreffig{fig4} shows, PrimNet takes the primal target image $\bm{x}^t_{p}$ and conditional image-label pairs $\{(\bm{x}^{c_j}_{p}, \bm{y}^{c_j}_{p})\}_{j=1}^m$ as input, and outputs the label of conditional supervision $\hat{\bm{z}}_{p}$, which is described in \zkreffig{fig5}.
	The output $\hat{\bm{z}}_{p}$ contains the intersection $\hat{\bm{z}}^{\cap}_{p}$ and the extra part $\hat{\bm{z}}^{\Delta}_{p}$ of the target label.
	Their gold standards are given by $\bm{z}^{\cap}_{p} = \bm{y}^t_{p}\odot \bm{y}^c_{p}$, and $\bm{z}^{\Delta}_{p} = \bm{y}^t_{p} \ominus (\bm{y}^t_{p}\odot \bm{y}^c_{p})$, respectively.
	The final segmentation result of $\bm{x}^t_p$ could be obtained by $\hat{\bm{y}}^t_p = \hat{\bm{z}}^\cap_p\oplus\hat{\bm{z}}^\Delta_p$.
	
	For DualNet, the input consists of the dual target image $\bm{x}^t_{d}$ and dual conditional image-label pairs $\{(\bm{x}^{c_j}_d, \bm{y}^{c_j}_d)\}_{j=1}^m$. 
	For selected class $c_s$, the target image and conditional image of PrimNet are swapped, represented as $\bm{x}^t_{d}=\bm{x}^{c_s}_p$ and $(\bm{x}^{c_s}_d,\bm{y}^{c_s}_d)=(\bm{x}^t_{p},\hat{\bm{y}}^{t_s}_p)$.
	For unselected label classes ($j\neq s$), dual condition images are the same with the primal conditional images, \textit{i.e}, $(\bm{x}^{c_j}_d,\bm{y}^{c_j}_d)=(\bm{x}^{c_j}_p,\bm{y}^{c_j}_p)$. 
	Let $\hat{\bm{z}}^t_{d}$ be the output of DualNet, which consists of the intersection $\hat{\bm{z}}^{\cap}_{d}$ and extra $\hat{\bm{z}}^{\Delta}_{d}$.
	Their gold standards are obtained by $\bm{z}^{\cap}_{d} = \bm{y}^t_{d}\odot \bm{y}^c_{d}$, and $\bm{z}^{\Delta}_{d} = \bm{y}^t_{d} \ominus (\bm{y}^t_{d}\odot\bm{y}^c_{d})$, respectively. 
	Similarly, the multi-label segmentation of $\bm{x}^c_d$ is given by $\hat{\bm{y}}^t_d = \hat{\bm{z}}^\cap_d\oplus\hat{\bm{z}}^\Delta_d$.
	
	Therefore, the final training objective of DCL is formulated, according to Eq.~(\ref{eq8}), as follows,
	\begin{equation}
		\mathcal{L} = (1-\lambda)\mathcal{L}_{cc}^{prim}(\hat{\bm{z}}_{p},\bm{z}_{p}, \bm{y}^c_{p}) + \lambda \mathcal{L}_{cc}^{dual}(\hat{\bm{z}}_{d},\bm{z}_{d}, \bm{y}^c_{d}),
		\label{eq7}
	\end{equation}
	where both $\mathcal{L}_{cc}^{prim}$ and $\mathcal{L}_{cc}^{dual}$ adopt the loss form in Eq.~(\ref{eq6}). Superscripts $prim$ and $dual$ denote the loss of PrimNet and DualNet, respectively.

	\begin{algorithm}[th]
		\small
		\caption{Deep Compatible Learning}
		\label{algo:DCL}
		\LinesNumbered 
		\KwIn{partially annotated dataset $\mathcal{S}_P$\; 
		}
		\KwOut{$\theta_p$ trained for image segmentation\;}
		\BlankLine
		Initialize PrimNet $\theta_p$ with random values.\\
		\While{$\theta_p$ not converged}{
			Sample target image $\bm{x}^t$ from $\mathcal{S}_P$\;
			Sample conditional images $\bm{x}^c(\{\bm{x}^{c_j}\}_{j=1}^m)$ from $\mathcal{S}_P$\;
			Compute conditional supervision $\bm{z}= g(\bm{y}^t, \bm{y}^c)$\;
			Update $\theta_p$ by minimizing Eq.(\ref{eq6})\;
		}
		Initialize $\theta_d$ of DualNet with pre-trained $\theta_p$\;
		\Repeat{$\theta_p$ and $\theta_d$ both converged}{
			Sample primal target image $\bm{x}^t_{p}$ from $\mathcal{S}_P$\;
			Sample primal conditional images $\bm{x}^c_{p}(\{\bm{x}^{c_j}_{p}\}_{j=1}^m)$ from $\mathcal{S}_P$\;
			Randomly select class $c_s$ from $\{c_j\}_{j=1}^m$\;
			Set selected primal conditional image $\bm{x}^{c_s}_{p}$ as dual target image $\bm{x}^t_{d}$\;
			Set $\bm{x}^c_{p}$ as dual conditional images $\bm{x}^c_{d}$, and replace $\bm{x}^{c_s}_{d}$ with $\bm{x}^t_{p}$\;
			Compute conditional supervision of PrimNet and DualNet:
			$\bm{z}_p= g(\bm{y}^t_p, \bm{y}_{p}^c), \bm{z}_d= g(\bm{y}^t_d, \bm{y}_{d}^c)$\;
			Update $\theta_p$ by minimizing Eq.(\ref{eq7})\;
			\If{$\theta_p$ converged}{
				$\theta_p \to \theta_d$\;
			}
		}
	\end{algorithm}
	\subsubsection{Training strategy}
	\label{sec3.3.3}
	In principle, the choice of label class $c_s$ used for dual compatibility can be arbitrary. However, if the partially annotated label class of primal target image is consistently selected for dual compatibility, the network would easily learn this label category, and only predicts the corresponding segmentation for the annotated label classes. This could lead to a failure for the training. Therefore, we use a more general method by randomly selecting label categories for dual compatibility.  
	As a result, each label category has the possibility of being selected to form a closed-loop. When there are enough training iterations, the dual compatibility can be achieved on all label classes. 
	
	In the conventional training process of dual learning, the primal network and the dual network are trained separately,  due to the difference between the primal task and the dual task. 
	Here, PrimNet and DualNet are both aimed to transfer the labels from the condition images to the target image, which allows for PrimNet and DualNet to adopt the same network architecture and share the same parameters.
	Therefore, we achieve this via an iterative method, as  Algorithm \ref{algo:DCL} shows.
	We first use the method described in \zkreffig{fig3} to train the PrimNet.
	Then, we initialize DualNet with the trained parameters of PrimNet.
	In the following iterations, we only keep the back-propagation process to update the model parameters of PrimNet,
	namely the gradient of $\mathcal{L}_{cc}^{prim}$ and $\mathcal{L}_{cc}^{dual}$ are only back-propagated to PrimNet.
	When the optimization of PrimNet is converged, we use the parameters of PrimNet to update that of DualNet. 
	Finally, in the inference stage, only the PrimNet is used for segmentation.

	\subsection{Deep Compatible Learning} 	\label{section3.4}
	We show that the proposed framework of DCL is generally applicable to existing loss functions. 
	We split a loss function into two parts, each for the positive label with ground truth $1$ and the negative label with ground truth $0$, respectively, based on which we formulate the $\mathcal{L}_P$ and $\mathcal{L}_N$ to form a compatible loss in Eq.(\ref{eq2}).  

	Then, $\mathcal{L}_N$ and $\mathcal{L}_P$ can be sequentially substituted into Eq.(\ref{eq2}), Eq.(\ref{eq4}), Eq.(\ref{eq5}), Eq.(\ref{eq6}), Eq.(\ref{eq7}) to obtain the final training objective $\mathcal{L}$.
	Here, we implement four loss functions for illustration, \textit{i.e.}, (1) binary cross-entropy loss, (2) Dice loss, (3) target adaptive loss and (4) exclusive loss. The former two are widely used in fully-supervised segmentation, while the latter two are mainly developed for partially-supervised segmentation.
	The derived formulations of $\mathcal{L}_P$ and $\mathcal{L}_N$ all satisfy the conditions in Eq.(3-4), which are discussed in Appendix C.
	
	\subsubsection{Application to existing loss functions for fully-supervised segmentation}
	
	\noindent\newline\textbf{Cross entropy loss:}
	The binary cross entropy loss function is formulated as $\mathcal{L}_{ce}(\hat{\bm{y}},\bm{y}) = -\sum_{i=1}^V\sum_{j=1}^m[y_{ij}\log(\hat{y}_{ij})+(1-y_{ij})\log(1-\hat{y}_{ij})]$.
	By requiring $y_{ij}=1$ and $y_{ij}=0$, we split the binary cross entropy into $\mathcal{L}_{P}^{[ce]}$ and $\mathcal{L}_{N}^{[ce]}$, as follows: 
	\begin{equation}
		\mathcal{L}_{P}^{[ce]}(\hat{\bm{y}},\bm{y}) =-\sum_{i=1}^V\sum_{j=1}^m\mathbbm{1}_{[y_{ij}=1]}\log(\hat{y}_{ij}), 
		\label{eq9}
	\end{equation}
	\begin{equation}
		\mathcal{L}_{N}^{[ce]}(\hat{\bm{y}},\bm{y}) =-\sum_{i=1}^V\sum_{j=1}^m\mathbbm{1}_{[y_{ij}=0]}\log(1-\hat{y}_{ij}).  
		\label{eq10}
	\end{equation}
	\noindent\newline\textbf{Dice loss:}
	Dice loss is written as $\mathcal{L}_{dsc}(\hat{\bm{y}},\bm{y})=\sum_{i=1}^V\sum_{j=1}^m(1-\frac{2y_{ij}\hat{y}_{ij}}{y_{ij}+\hat{y}_{ij}})$.
	Note that in the binary classification, Dice loss is only calculated for the case of $y_{ij}=1$. 
	Therefore, it is equivalent to the positive component of loss function. 
	For negative labels with $y_{ij} = 0$, by replacing $y_{ij}$ with ($1-y_{ij}$), we generalize Dice loss to the negative label. Therefore, $\mathcal{L}_P^{[d]}$ and $\mathcal{L}_N^{[d]}$ are given by,
	\begin{equation}
		\mathcal{L}_{P}^{[d]}(\hat{\bm{y}},\bm{y}) = 
		\sum_{i=1}^V\sum_{j=1}^m\mathbbm{1}_{[y_{ij}=1]}\left(1-\frac{2\hat{y}_{ij}}{1+\hat{y}_{ij}}\right),
		\label{eq11}
	\end{equation}
	\begin{equation}
		\mathcal{L}_{N}^{[d]}(\hat{\bm{y}},\bm{y}) =
		\sum_{i=1}^V\sum_{j=1}^m\mathbbm{1}_{[y_{ij}=0]}\left(\frac{2\hat{y}_{ij}}{1+\hat{y}_{ij}}\right). 
		\label{eq12}
	\end{equation}
	
	\subsubsection{Application to existing loss functions for partially-supervised segmentation}
	
	The target adaptive loss~\cite{fang2020multi} and exclusive loss~\cite{shi2020marginal} were specifically designed for partially-supervised segmentation. 
	The former calculates the loss for positive labels by exploiting the merged marginal probability,
	while the latter penalizes the prediction errors of negative labels by requiring that different structures do not cross each other.  
	\textit{Therefore, here we solely develop the positive loss $\mathcal{L}_P$ from the target adaptive loss, and formulate the negative $\mathcal{L}_N$ from the exclusive loss.} 
	
	\noindent\newline\textbf{Target adaptive loss:}
	The target adaptive loss ($\mathcal{L}_{ta}$) is proposed to merge all unlabeled categories with the background label.  
	Therefore, $\mathcal{L}_{ta}$ calculates the loss for two cases; one is when the ground truth is 1 ($y_{ij} = 1$), and the other is when the merged background label equals 1 ($\sum_{j\in\bar{\bm{c}}_q} y_{ij}=1$).
	In both cases, the loss is calculated for the positive label (or merged label).
	Then, it is obtained that $\mathcal{L}_P=\mathcal{L}_{ta}$.
	Here, we give two forms of $\mathcal{L}_{P}^{[ta]}$ based on cross entropy and Dice loss, respectively denoted as
	$\mathcal{L}^{[ta,ce]}_{P}$ and $\mathcal{L}_{P}^{[ta,d]}$:
	\begin{equation} 
		\begin{aligned}
			\mathcal{L}^{[ta,ce]}_{P}(\hat{\bm{y}},\bm{y}) =& - \sum_{i=1}^V\sum_{j=1}^m[\mathbbm{1}_{[y_{ij}=1]}\log \hat{y}_{ij}\\
			&-\mathbbm{1}_{[\sum_{j\in\bar{\bm{c}}_q} y_{ij}=1]}\log(\sum_{j\in\bar{\bm{c}}_q} \hat{y}_{ij})],
		\end{aligned}
		\label{eq15}
	\end{equation}
	\begin{equation}
		\small
		\begin{aligned}
			\mathcal{L}^{[ta,d]}_{P}(\hat{\bm{y}},\bm{y}) =& \sum_{i=1}^V\sum_{j=1}^m[\mathbbm{1}_{[y_{ij}=1]}(1-\frac{2\hat{y}_{ij}}{\hat{y}_{ij}+1})  \\
			&+\mathbbm{1}_{[\sum_{j\in\bar{\bm{c}}_q} y_{ij}=1]}\left(1-\frac{2 \sum_{j\in\bar{\bm{c}}_q}\hat{y}_{ij}}{\sum_{j\in\bar{\bm{c}}_q}\hat{y}_{ij}+1}\right)].
		\end{aligned}
		\label{eq16}
	\end{equation}
	
	\noindent\newline\textbf{Exclusive loss:} 
	Exclusive loss ($\mathcal{L}_{ex}$) considers the exclusivity between different structures and applies the exclusiveness as prior knowledge to each image pixel. Complementary to $\mathcal{L}_{ta}$, the exclusive loss is used in two cases: the ground truth equals 0 ($y_{ij}=0$), or the merged background label value is 0 ($\sum_{j\in\bar{\bm{c}}_q} y_{ij}=0$). In other words, exclusive loss calculates the loss for negative labels (or merged labels). Similarly, we formulate two forms of $\mathcal{L}_N^{[ex]}$ based on cross entropy and Dice loss,
	respectively denoted as $\mathcal{L}^{[ex,ce]}_{N}$ and $\mathcal{L}^{[ex,d]}_{N}$, \textit{i.e.},
	\begin{equation}
		\begin{aligned}
			\mathcal{L}^{[ex,ce]}_{N}(\hat{\bm{y}},\bm{y}) =& \sum_{i=1}^V\sum_{j=1}^m[\mathbbm{1}_{[y_{ij}=0]}\log (\hat{y}_{ij}+\epsilon)\\
			&+\mathbbm{1}_{[\sum_{j\in\bar{\bm{c}}_q} y_{ij}=0]}\log (\sum_{j\in\bm{c}_q} \hat{y}_{ij}+\epsilon)],
		\end{aligned}
	\end{equation}
	\begin{equation}
		\begin{aligned}
			\mathcal{L}^{[ex,d]}_{N}(\hat{\bm{y}},\bm{y})=&\sum_{i=1}^V\sum_{j=1}^m[\mathbbm{1}_{[y_{ij}=0]}\frac{2\hat{y}_{ij}}{\hat{y}_{ij}+1}\\&+\mathbbm{1}_{[\sum_{j\in\bar{\bm{c}}_q} y_{ij}=0]}\frac{2\sum_{j\in\bm{c}_q} \hat{y}_{ij}}{\sum_{j\in\bm{c}_q} \hat{y}_{ij}+1}],
		\end{aligned}
	\end{equation}
	where $\epsilon$ is set to be 1. 
	
	\section{Experiments and Results}
	\label{section5}
	In the experiments, we first validated the compatibility, data insensitivity and applicability in Section \ref{sec4.2}. 
	Then, the performance of DCL was evaluated on three image segmentation tasks with comparisons with state-of-the-art methods, in Section \ref{sec4.3}.
	Note that although using a combination of several loss functions could achieve slightly better results, here we mainly used the cross entropy loss to illustrate the performance on the three segmentation tasks for efficiency. 
	The applicability study in Section~\ref{sec4.2.3} demonstrated that DCL with different loss functions and combinations only has marginal difference on the performance.
	
	\subsection{Experiments Setup}
	\subsubsection{Task and datasets.}
	We validated the proposed deep compatible framework on three segmentation tasks, \textit{i.e.}, ACDC \cite{bernard2018deep}, MSCMRseg \cite{zhuang2019multivariate,zhuang2016multivariate} and MMWHS \cite{zhuang2016multi,zhuang2019evaluation}:
	\begin{itemize}
		\item \textbf{ACDC}\cite{bernard2018deep}. The MICCAI'17 Automatic Cardiac Diagnosis Challenge dataset comprises of short-axis cardiac cine-MRIs of 100 patients from 5 groups: one group with 20 healthy controls and 20 subjects in each remaining group with four different abnormalities. Manual segmentation results are provided for the end-diastolic (ED) and end-systolic (ES) cardiac phases for left ventricle (LV), right ventricle (RV) and myocardium (MYO). We randomly divided the 100 subjects into 60 training subjects, 20 validation subjects and 20 test subjects.
		\item \textbf{MSCMRseg}\cite{zhuang2016multivariate,zhuang2019multivariate}. The MICCAI'19 Multi-sequence Cardiac MR Segmentation Challenge dataset consists of 45 multi-sequence CMR images from patients who underwent cardiomyopathy and the goal is to achieve automatic segmentation of LV, RV, MYO from LGE CMR.
		In this study, we used the 45 LGE images and divided them into 3 sets of 25 (training), 5 (validation), and 15 (test) images for experiments. 
		\item \textbf{MMWHS}\cite{zhuang2016multi,zhuang2019evaluation}. The MICCAI'17 Multi-Modality Whole Heart Segmentation challenge provides 60 cardiac MRI images. Segmentation masks of whole heart substructures include the left ventricle (LV), right ventricle (RV), left atrium (LA), right atrium (RA), myocardium (Myo), ascending aorta (AO), and the pulmonary artery (PA). Here, we divided the 60 MRI images into 3 sets of 20 (training), 10 (validation), and 30 (test) images for experiments.
	\end{itemize}
	For the partially-supervised learning experiments conducted on the ACDC and MSCMRseg datasets, we only used one randomly selected label of an image for training leaving all the others unlabeled, to simulate a partially annotated case. 
	Similarly, for partially-supervised segmentation experiments on MMWHS dataset, we used two randomly selected labels of each image, which account for $1/4$ of the total label categories.
	The number of annotated images for each label class was set to be equal.
	
	\subsubsection{Training procedure.}
	For ACDC and MSCMRseg datasets, we adopted the 2D UNet in \cite{baumgartner2017exploration}, referred to UNet$^+$ as the backbone for all segmentation networks, which is a variant of UNet \cite{ronneberger2015u}. 
	For MMWHS dataset, we trained the 3D nnUNet\cite{isensee2021nnu} for all segmentation networks. The nnUNet is known to be the state-of-the-art network for 3D segmentation tasks. 
	
	For experiments on ACDC and MSCMRseg datasets, the in-plane resolution of all these datasets was resampled to 1.37$\times$1.37mm. We then cropped or padded the images to a fixed size of 212$\times$212 and normalized the intensity of each image to zero mean and unit variance. We used the same augmentation method of rotation, flip, and contrast adjustment for all the experiments implemented. For experiments on MMWHS dataset, we did not use any preprocessing as nnUNet performs the data pre-processing automatically.
	
	The DCL was trained in a two-stage manner, as illustrated in Algorithm \ref{algo:DCL}. 
	In the first stage, only the PrimNet was trained to minimize the conditional compatibility loss $\mathcal{L}_{cc}$ in Eq.(\ref{eq6}). In the second stage, the PrimNet and DualNet were trained jointly to minimize $\mathcal{L}$ in Eq.(\ref{eq7}).
	PrimNet was updated through backpropagation. DualNet was updated to the optimal weight of PrimNet, which was evaluated on the validation set.
	Note that only PrimNet was used to segment a target image in the test (inference) phase. 
	
	
	\begin{table*}[!thb]
		\caption{Ablation study: DCL for image segmentation with different settings, including loss functions, the use of conditional images ($\bm{x}^c$) and DualNet (dual). From model \#1 to \#6, the Dice scores of the method gradually increase, with * indicating statistically significant improvement given by a Wilcoxon signed-rank test (p$<$0.05). \textbf{Bold} denotes the best results. Please refer to the text for details.}\label{tab7}
		\begin{center}
			\resizebox{1\textwidth}{!}{
				\begin{tabular}{cccccccccccccc}
					\hline
					\multirow{2}{*}{Model}&\multirow{2}{*}{$\mathcal{L}_{ce}$}&\multirow{2}{*}{$\mathcal{L}^{[ce]}_{P}$}&\multirow{2}{*}{$\mathcal{L}_{\text{comp}}$}&\multirow{2}{*}{$\bm{x}^c$}&\multirow{2}{*}{$\mathcal{L}_{\text{prior}}$}&\multirow{2}{*}{dual}&\multicolumn{3}{c}{ED} &\multicolumn{3}{c}{ES}&\multirow{2}{*}{Avg}\\  
					\cmidrule(lr){8-10}\cmidrule(lr){11-13}
					&&&&&&& LV & MYO & RV & LV & MYO & RV& \\
					\hline
					\multicolumn{1}{c|}{\#1}&$\checkmark$&$\times$&$\times$&$\times$&$\times$&\multicolumn{1}{c|}{$\times$}&.767$\pm$.198&.740$\pm$.082&\multicolumn{1}{c|}{.199$\pm$.224}&.651$\pm$.215&.722$\pm$.108&.115$\pm$.196&\multicolumn{1}{|c}{.534}\\
					\multicolumn{1}{c|}{\#2}&$\times$&$\checkmark$&$\times$&$\times$&$\times$&\multicolumn{1}{c|}{$\times$}&.939$\pm$.047$^*$&.813$\pm$.046$^*$&\multicolumn{1}{c|}{.742$\pm$.185$^*$}&.841$\pm$.140$^*$&.786$\pm$.078$^*$&.532$\pm$.232$^*$&\multicolumn{1}{|c}{.778$^*$}\\
					\multicolumn{1}{c|}{\#3}&$\times$&$\times$&$\checkmark$&$\times$&$\times$&\multicolumn{1}{c|}{$\times$}&.952$\pm$.027&.844$\pm$.040$^*$&\multicolumn{1}{c|}{.792$\pm$.124}&.879$\pm$.110$^*$&.814$\pm$.078&.589$\pm$.186&\multicolumn{1}{|c}{.814$^*$}\\
					\multicolumn{1}{c|}{\#4}&$\times$&$\times$&$\checkmark$&$\checkmark$&$\times$&\multicolumn{1}{c|}{$\times$}&.950$\pm$.031&.852$\pm$.031&\multicolumn{1}{c|}{.852$\pm$.091$^*$}&.863$\pm$.134&.842$\pm$.056$^*$&.661$\pm$.167$^*$&\multicolumn{1}{|c}{.839$^*$}\\
					\multicolumn{1}{c|}{\#5}&$\times$&$\times$&$\checkmark$&$\checkmark$&$\checkmark$&\multicolumn{1}{c|}{$\times$}&.952$\pm$.035$^*$&$\bm{.872}\pm$.041$^*$&\multicolumn{1}{c|}{.894$\pm$.070$^*$}&.861$\pm$.156&$\bm{.865}\pm$.063$^*$&.771$\pm$.146$^*$&\multicolumn{1}{|c}{.871$^*$}\\
					\multicolumn{1}{c|}{\#6}&$\times$&$\times$&$\checkmark$&$\checkmark$&$\checkmark$&\multicolumn{1}{c|}{$\checkmark$}&$\bm{.956}\pm$.025&$\bm{.872}\pm$.040&\multicolumn{1}{c|}{$\bm{.895}\pm$.068}&$\bm{.873}\pm$.138&.864$\pm$.067&$\bm{.794}\pm$.119$^*$&\multicolumn{1}{|c}{$\bm{.877}^*$}\\
					\hline
			\end{tabular}}
		\end{center}
	\end{table*}  
	
	\subsubsection{Implementation.} The proposed framework was implemented using Keras library with TensorFlow 2.0 backend. We trained our network using Adam optimizer with the initial learning rate of $10^{-2}$ in first stage, and $10^{-4}$ in second stage. The $\lambda$ in Eq.(\ref{eq7}) was empirically set to be $0.2$ and remained the same for all datasets. All the experiments were run on a Nvidia RTX 2080Ti GPU.
	
	\subsubsection{Baselines.}
	The segmentation method using the proposed framework was denoted as DCL.
	For comparisons, we implemented a number of other algorithms.  
	
	Our main partially-supervised baseline was the method proposed by Shi \textit{et al.}\cite{shi2020marginal}, denoted as MMEE, which combined four forms from the target adaptive loss and exclusive loss based on cross entropy and Dice loss in Eq.(15-18), respectively.
	We also compared to UNet$^+$ network trained on fully annotated images, denoted as UNet$^+_F$, with different numbers of training images to simulate similar amount of label supervision.
	Besides, we implemented the conventional CE loss and UNet, which ignored the loss of the unlabeled pixels during the back-propagation. This model was trained with CE-based positive loss $\mathcal{L}^{[ce]}_P$ on the same set of single-label images as DCL, and was denoted as UNet$^+_P$.
	
	For comparisons with fully-supervised baseline methods, we included the segmentation results reported in the literature for the three segmentation tasks for reference. 
	In the case of full annotation, we extended DCL by considering a fully-labeled training image with four labels as four single-label images, and a training image with eight labels as four dual-label images.
	The details will be given in the following studies.

	\subsection{Parameter Studies}
	This section consists of three studies, including the ablation study, data sensitivity study, and application study. They are respectively designed to verify the compatibility, data insensitivity and applicability of the proposed DCL framework.
	\label{sec4.2}
	
	\subsubsection{Ablation study}
	
	In this study, we show that compatibility can improve the performance of the model. 
	The study consists of six experiments for partially-supervised segmentation trained on 60 single-label subjects using the ACDC dataset.
	The parameters include loss functions, the use of conditional images ($\bm{x}^c$) and DualNet ($dual$). 
	The settings of loss functions include the conventional cross entropy ($\mathcal{L}_{ce}$),  positive loss based on cross entropy ($\mathcal{L}^{[ce]}_P$) in Eq.(\ref{eq9}), compatible loss ($\mathcal{L}_{comp}$) in Eq.(\ref{eq4}), prior loss ($\mathcal{L}_{prior}$) in Eq.(\ref{eq5}).
	The results of our ablation study and related parameter settings are summarized in Table~\ref{tab7}.
	We have the following three major observations from this study: the effects of compatibility, conditional compatibility and dual compatibility.
	
	\textbf{The effect of compatibility:} 
	The experiment results obtained by the models trained with compatible loss ($\mathcal{L}_P^{[ce]}$, $\mathcal{L}_{\text{comp}}$, $\mathcal{L}_\text{prior}$) are generally better than those by the model with incompatible loss ($\mathcal{L}_{ce}$).
	Model \#1 with conventional CE loss $\mathcal{L}_{ce}$ suffered from incompatibility, as described in Proposition 1.
	Model \#2 with CE-based positive loss $\mathcal{L}^{[ce]}_P$, performed significantly better than the model \#1 with conventional CE loss $\mathcal{L}_{ce}$; and the average Dice score jumped from $53.4\%$ to $77.8\%$. 
	When using compatible loss $\mathcal{L}_{comp}$, model \#3 obtained a significant gain of $3.6\%$ in Dice, from $77.8\%$ to $81.4\%$, compared to model \#2 with $\mathcal{L}^{[ce]}_P$.
	This benefit was even more evident on the challenging RV segmentation, where increases of Dice by $5.0\%$ and $5.7\%$ were seen respectively on ED and ES phases.
	
	\textbf{The effect of conditional compatibility:} When using conditional images for label propagation, the model \#4 performed significantly better than model \#3, with an increase of $2.5\%$ of average Dice ($0.839$ vs. $0.814$).
	When conditional compatibility was introduced by $\mathcal{L}_{prior}$, the average Dice score of model \#5 further jumped from $83.9\%$ to $87.1\%$, with $11\%$ Dice score increase of RV in ES phases, demonstrating the effectiveness of the conditional compatibility for the challenging tasks.
	
	\textbf{The effect of dual compatibility:} For model \#6, one could also observe a statistically significant gain of average Dice of $0.6\%$ using DualNet and a $2.3\%$ increase on the RV segmentation of ES phases. This was attributed to the fact that the dual closed-loop setting can reduce the generalization error bound and provide substantial supervision for unlabeled structures, as discussed in Section~\ref{section3.3}.
	
	\subsubsection{Data sensitivity study}
	\label{section4.4} 
	\begin{table*}[tb]
		\caption{Data sensitivity: 6 sets of experiments using different amounts of fully annotated and single annotated training images. Full:part indicates the number of fully annotated subjects and single annotated subjects in trainSet. The total number of annotated labels are given in the column of \textit{Total}. As the proportion of fully labeled subjects increases, the general performance of DCL improves and tends to converge, with * indicating statistically significant improvement given by a Wilcoxon signed-rank test ,\textit{i.e.}, p$<$0.05.}\label{tab8}
		\begin{center}
			\resizebox{0.9\linewidth}{!}{
				\begin{tabular}{cccccccccc}
					\hline
					\multirow{2}{*}{Method}&\multirow{2}{*}{Full:part} &\multirow{2}{*}{Total}&\multicolumn{3}{c}{ED} &\multicolumn{3}{c}{ES}&\multirow{2}{*}{Avg}\\  
					\cmidrule(lr){4-6}\cmidrule(lr){7-9}
					&&& LV & MYO & RV & LV & MYO & RV& \\
					\hline
					\multirow{6}{*}{DCL}&\multicolumn{1}{|c|}{00:60}&\multicolumn{1}{c|}{60}&.956$\pm$.025&.872$\pm$.040&\multicolumn{1}{c|}{.895$\pm$.068}&.873$\pm$.138&.864$\pm$.067&.794$\pm$.119&\multicolumn{1}{|c}{.877}\\
					&\multicolumn{1}{|c|}{12:48}&\multicolumn{1}{c|}{96}&.965$\pm$.012$^*$&.876$\pm$.032&\multicolumn{1}{c|}{.924$\pm$.036$^*$}&.905$\pm$.105$^*$&.883$\pm$.040&.791$\pm$.170&\multicolumn{1}{|c}{.892$^*$}\\
					&\multicolumn{1}{|c|}{24:36}&\multicolumn{1}{c|}{132}&$\bm{.968}\pm$.012$^*$&$\bm{.895}\pm$.024$^*$&\multicolumn{1}{c|}{$\bm{.922}\pm$.041}&.912$\pm$.090$^*$&.897$\pm$.029$^*$&.810$\pm$.134$^*$&\multicolumn{1}{|c}{.902$^*$}\\
					&\multicolumn{1}{|c|}{36:24}&\multicolumn{1}{c|}{168}&.962$\pm$.025&.885$\pm$.025&\multicolumn{1}{c|}{.912$\pm$.061}&.914$\pm$.079&.895$\pm$.025&.808$\pm$.145&\multicolumn{1}{|c}{.897}\\
					&\multicolumn{1}{|c|}{48:12}&\multicolumn{1}{c|}{204}&.965$\pm$.017$^*$&.892$\pm$.025$^*$&\multicolumn{1}{c|}{$\bm{.922}\pm$.040}&$\bm{.927}\pm$.070$^*$&$\bm{.900}\pm$.031$^*$&.815$\pm$.193&\multicolumn{1}{|c}{.905$^*$}\\
					&\multicolumn{1}{|c|}{60:00}&\multicolumn{1}{c|}{240}&.962$\pm$.028&.891$\pm$.025&\multicolumn{1}{c|}{.920$\pm$.055}&.917$\pm$.082&.898$\pm$.030&$\bm{.858}\pm$.082&\multicolumn{1}{|c}{$\bm{.909}$}\\
					\hline
					\multicolumn{1}{c|}{UNet$^+_F$}&\multicolumn{1}{c|}{60:00}&\multicolumn{1}{c|}{240}&.961$\pm$.016&.881$\pm$.026&\multicolumn{1}{c|}{.901$\pm$.063}&.908$\pm$.063&.888$\pm$.025&.786$\pm$.093&\multicolumn{1}{|c}{.889}\\
					\hline
			\end{tabular}}
		\end{center}
	\end{table*} 
	
	\begin{table*}[tb]
		\caption{The performance on ACDC dataset of DCL with different basis of loss functions for partially-supervised segmentation. CE denotes cross entropy loss. Dice implies Dice loss. The MMEE represents the combination of target adaptive loss and exclusive loss based on cross entropy and Dice loss. We use Wilcoxon signed-rank test (p$<$0.05) for significance test. The superscript $^*$ denotes the statistically significant improvement of DCL for the same loss function combination.} 
		\label{tab5}
		\begin{center}
			\resizebox{0.9\textwidth}{!}{
				\begin{tabular}{lcccccccc}
					\hline
					\multirow{2}{*}{Methods}&\multirow{2}{*}{Total} & \multicolumn{3}{c}{ED} &\multicolumn{3}{c}{ES}&\multirow{2}{*}{Avg}\\  
					\cmidrule(lr){3-5}\cmidrule(lr){6-8}
					& &LV & MYO & RV & LV & MYO & RV& \\
					\hline
					\multicolumn{9}{l}{with DCL}\\
					\hline
					\multicolumn{1}{l|}{CE}&\multicolumn{1}{c|}{60$\times$1}&.956$\pm$.025$^*$&.872$\pm$.040$^*$&\multicolumn{1}{c|}{.895$\pm$.068$^*$}&.873$\pm$.138$^*$&.864$\pm$.067$^*$&.794$\pm$.119$^*$&\multicolumn{1}{|c}{.877$^*$}\\
					\multicolumn{1}{l|}{CE+Dice}&\multicolumn{1}{c|}{60$\times$1}&$\bm{.962}\pm$.018$^{*}$&.880$\pm$.026$^*$&\multicolumn{1}{c|}{$\bm{.908}\pm$.058$^*$}&$\bm{.910}\pm$.018$^{*}$&$\bm{.891}\pm$.026$^{*}$ & .780$\pm$.058$^{*}$ &\multicolumn{1}{|c}{.890$^{*}$}\\
					\multicolumn{1}{l|}{MMEE}&\multicolumn{1}{c|}{60$\times$1}&$\bm{.962}\pm$.018&.883$\pm$.026&\multicolumn{1}{c|}{.905$\pm$.055}&.903$\pm$.094&.889$\pm$.029&$\bm{.810}\pm$.107$^{*}$&\multicolumn{1}{|c}{$\bm{.893}$}\\
					\hline
					\multicolumn{9}{l}{without DCL}\\
					\hline
					\multicolumn{1}{l|}{CE}&\multicolumn{1}{c|}{60$\times$1}&.939$\pm$.047&.813$\pm$.046&\multicolumn{1}{c|}{.742$\pm$.185}&.841$\pm$.140&.786$\pm$.078&.532$\pm$.232&\multicolumn{1}{|c}{.778}\\
					\multicolumn{1}{l|}{CE+Dice}&\multicolumn{1}{c|}{60$\times$1}&.947$\pm$.027&.832$\pm$.059&\multicolumn{1}{c|}{.790$\pm$.140}&.846$\pm$.114&.813$\pm$.054&.598$\pm$.196&\multicolumn{1}{|c}{.807}\\
					\multicolumn{1}{l|}{MMEE}&\multicolumn{1}{c|}{60$\times$1}&.959$\pm$.030&$\bm{.885}\pm$.032&\multicolumn{1}{c|}{$\bm{.908}\pm$.066}&.892$\pm$.107&.883$\pm$.029&.768$\pm$.149&\multicolumn{1}{|c}{.884}\\
					\hline
			\end{tabular}}
		\end{center}
	\end{table*}

	This study verifies the robustness of DCL with different amounts of supervision. Six DCL models were trained using different proportions between fully annotated and single annotated training images. 
	Table~\ref{tab8} shows that the average Dice scores of DCL on the ACDC dataset.
	DCL generally improved the performance (mean Dice scores) as the fully annotated training images and total amount of annotations increase.
	Interestingly, when the total number reaches 96, DCL had demonstrated slightly better average Dice ($0.892$ vs $0.889$) than the fully-supervised UNet (UNet$^+_F$) trained on all data, \textit{i.e.}, $60\times 4 =240$ labels. This further confirms the effectiveness and efficiency of the proposed compatible framework. 
	Furthermore, one can observe from Table~\ref{tab8} that the general performance of DCL converged when the Total labels reached 132. However, the Dice of RV segmentation on ES phases consistently increased with respect to the increased amount of supervisions, and DCL gained an evidently better ES-RV Dice score when the Total supervision was maximized from the available training images, \textit{i.e.},  full:part = 60:00 and Total = 240. 
	This was probably due to the fact that the ES-RV segmentation was more challenging than the other structures, and more supervision was needed from the limited training data. 
	Finally, we observed that DCL demonstrated a small fluctuation in Dice scores when 
	Total=168 (full:part=36:24) compared to that of Total=132 (full:part=24:36). This may be due to the random effects of the division of the training images.

	\subsubsection{Application studies with different loss functions}
	\label{sec4.2.3}
	In this study, we conducted experiments on two segmentation tasks to measure the impact of DCL when it was applied to different loss functions. Three combinations of loss functions were used as benchmarks: conventional cross entropy loss (CE); a combination of cross entropy and dice loss (CE + Dice); a loss combining four forms from the target adaptive loss and exclusive loss respectively based on cross entropy and dice loss (MMEE).
	We plugged these loss functions into the proposed DCL framework and implemented their compatible forms, which are described in detail in Section \ref{section3.4}.
	
	\noindent\newline\textbf{ACDC Dataset:} Table~\ref{tab5} shows the performance of the conventional benchmarks with and without DCL on single annotated ACDC dataset. DCL significantly improved the segmentation performance (mean Dice scores) for all the three combinations of loss functions.
	For CE and CE+Dice, the average dice scores of DCL increased by $10\%$ and $8.3\%$, respectively.
	For MMEE, DCL achieved a convincing improvement of $4.2\%$  ($0.810$ vs $0.768$) on RV of ES phases and a marginal improvement of $0.9\%$ ($0.893$ vs $0.884$) average Dice score. 
	Additionally, we found that when DCL was used, the performance difference between different basis of loss functions became less evident. 
	With DCL, the performance of MMEE performed slightly better than CE, from $87.7\%$ to $89.3\%$, with an increase of $1.6\%$ in Dice. 
	However, when no DCL was applied, MMEE achieved much better Dice score than CE ($0.884$ vs $0.778$). 
	This is probably because MMEE was specifically designed for partially-supervised learning.
	These results further prove the general applicability and robustness of the proposed DCL.
	
	\noindent\newline\textbf{MSCMR Dataset:} 
	To further study the effect of DCL with limited training data, we conducted additional six sets of experiments on the MSCMRseg dataset, which contains only 25 training images. In this more challenging task, DCL demonstrated a more significant impact. 

\begin{table}[htb]
	\caption{The performance on MSCMR dataset of DCL with different basis of loss functions for partially-supervised segmentation.}\label{tab6}
	\begin{center}
		\resizebox{1\linewidth}{!}{
			\begin{tabular}{lccccc}
				\hline
				Methods& Total& LV & MYO & RV &\multicolumn{1}{|c}{Avg}\\
				\hline
				\multicolumn{6}{l}{with DCL}\\
				\hline
				\multicolumn{1}{l|}{CE}&\multicolumn{1}{c|}{25$\times$1}&.893$\pm$.048$^*$&.774$\pm$.049$^*$&$\bm{.769}\pm$.140$^*$&\multicolumn{1}{|c}{.812$^*$}\\
				\multicolumn{1}{l|}{CE+Dice}&\multicolumn{1}{c|}{25$\times$1}&.880$\pm$.060$^*$&.773$\pm$.068$^*$&.703$\pm$.199$^*$&\multicolumn{1}{|c}{.785$^{*}$}\\
				\multicolumn{1}{l|}{MMEE}&\multicolumn{1}{|c|}{25$\times$1}&$\bm{.897}\pm$.034&$\bm{.786}\pm$.050$^*$&$\bm{.769}\pm$.101$^*$&\multicolumn{1}{|c}{$\bm{.817}^{*}$}\\ 
				\hline
				\multicolumn{6}{l}{without DCL}\\
				\hline
				\multicolumn{1}{l|}{CE}&\multicolumn{1}{c|}{25$\times$1}&.658$\pm$.217&.556$\pm$.081&.262$\pm$.102&\multicolumn{1}{|c}{.492}\\
				\multicolumn{1}{l|}{CE+Dice}&\multicolumn{1}{c|}{25$\times$1}&.518$\pm$.241&.590$\pm$.080&.531$\pm$.140&\multicolumn{1}{|c}{.546}\\
				\multicolumn{1}{l|}{MMEE}&\multicolumn{1}{c|}{25$\times$1}&.889$\pm$.059&.750$\pm$.092&.667$\pm$.092&\multicolumn{1}{|c}{.769}\\
				\hline
		\end{tabular}}
	\end{center}
\end{table}

	\begin{table*}[t]
		\caption{The performance (Dice scores) on ACDC dataset for partially- and fully- supervised segmentation.  Column \textit{Total} indicates the total number of annotated labels in training images. $\text{Superscript}^{\dag}$ denotes the model was trained on 100 annotated subjects. }
		\scriptsize
		\begin{center}
			\resizebox{0.9\textwidth}{!}{
				\begin{tabular}{lcccccccc}
					\hline
					\multirow{2}{*}{Methods}&\multirow{2}{*}{Total} & \multicolumn{3}{c}{ED} &\multicolumn{3}{c}{ES}&\multirow{2}{*}{Avg}\\  
					\cmidrule(lr){3-5}\cmidrule(lr){6-8}
					& &LV & MYO & RV & LV & MYO & RV& \\
					\hline
					\multicolumn{9}{l}{partial annotation supervision}\\
					\hline
					\multicolumn{1}{l|}{UNet$^+_P$}&\multicolumn{1}{c|}{60$\times$1}&.939$\pm$.047&.813$\pm$.046&\multicolumn{1}{c|}{.742$\pm$.185}&.841$\pm$.140&.786$\pm$.078&.532$\pm$.232&\multicolumn{1}{|c}{.778}\\
					\multicolumn{1}{l|}{UNet$^+_F$}&\multicolumn{1}{c|}{15$\times$4}&.925$\pm$.066&.840$\pm$.067&\multicolumn{1}{c|}{.811$\pm$.094}&.849$\pm$.126&.852$\pm$.053& .721$\pm$.157 & \multicolumn{1}{|c}{.834}\\
					\multicolumn{1}{l|}{DCL$_{\text{ce}}$}&\multicolumn{1}{c|}{60$\times$1}&.956$\pm$.025&.872$\pm$.040&\multicolumn{1}{c|}{.895$\pm$.068}&.873$\pm$.138&.864$\pm$.067&.794$\pm$.119&\multicolumn{1}{|c}{.877}\\
					\multicolumn{1}{l|}{MMEE\cite{shi2020marginal}}&\multicolumn{1}{c|}{60$\times$1}&.959$\pm$.030&$\bm{.885}\pm$.032&\multicolumn{1}{c|}{$\bm{.908}\pm$.066}&.892$\pm$.107&.883$\pm$.029&.768$\pm$.149&\multicolumn{1}{|c}{.884}\\
					\multicolumn{1}{l|}{DCL$_{\text{MMEE}}$}&\multicolumn{1}{c|}{60$\times$1}&$\bm{.962}\pm$.018&.883$\pm$.026&\multicolumn{1}{c|}{.905$\pm$.055}&$\bm{.903}\pm$.094&$\bm{.889}\pm$.029&$\bm{.810}\pm$.107&\multicolumn{1}{|c}{$\bm{.893}$}\\
					\hline
					\multicolumn{9}{l}{full annotation supervision}\\
					\hline
					\multicolumn{1}{l|}{Avg\cite{bernard2018deep}$^{\dag}$}&\multicolumn{1}{c|}{100$\times 4$}&.950&.873&\multicolumn{1}{c|}{.909}&.893&.886&.848&\multicolumn{1}{|c}{.893}\\
					\multicolumn{1}{l|}{UNet$^+_F$\cite{baumgartner2017exploration}$^\dag$}&\multicolumn{1}{c|}{100$\times$4}& .963 &.892&\multicolumn{1}{c|}{.932}&.911&.901&.883&\multicolumn{1}{|c}{.914}\\
					\multicolumn{1}{l|}{UNet$^+_F$}&\multicolumn{1}{c|}{60$\times$4}&.961$\pm$.016&.881$\pm$.026&\multicolumn{1}{c|}{.901$\pm$.063}& .908$\pm$.063&.888$\pm$.025&.786$\pm$.093&\multicolumn{1}{|c}{.889}\\
					\multicolumn{1}{l|}{DCL$_{ce}$}&\multicolumn{1}{c|}{60$\times$4}&.962$\pm$.028&.891$\pm$.025&\multicolumn{1}{c|}{.920$\pm$.055}&.917$\pm$.082&.898$\pm$.030&.858$\pm$.082&\multicolumn{1}{|c}{.909}\\
					\hline
				\end{tabular}
			}
		\end{center} 
		\label{tab2}\end{table*}
	
	\begin{figure*}[thb]
		\centering
		\includegraphics[width=0.85\textwidth]{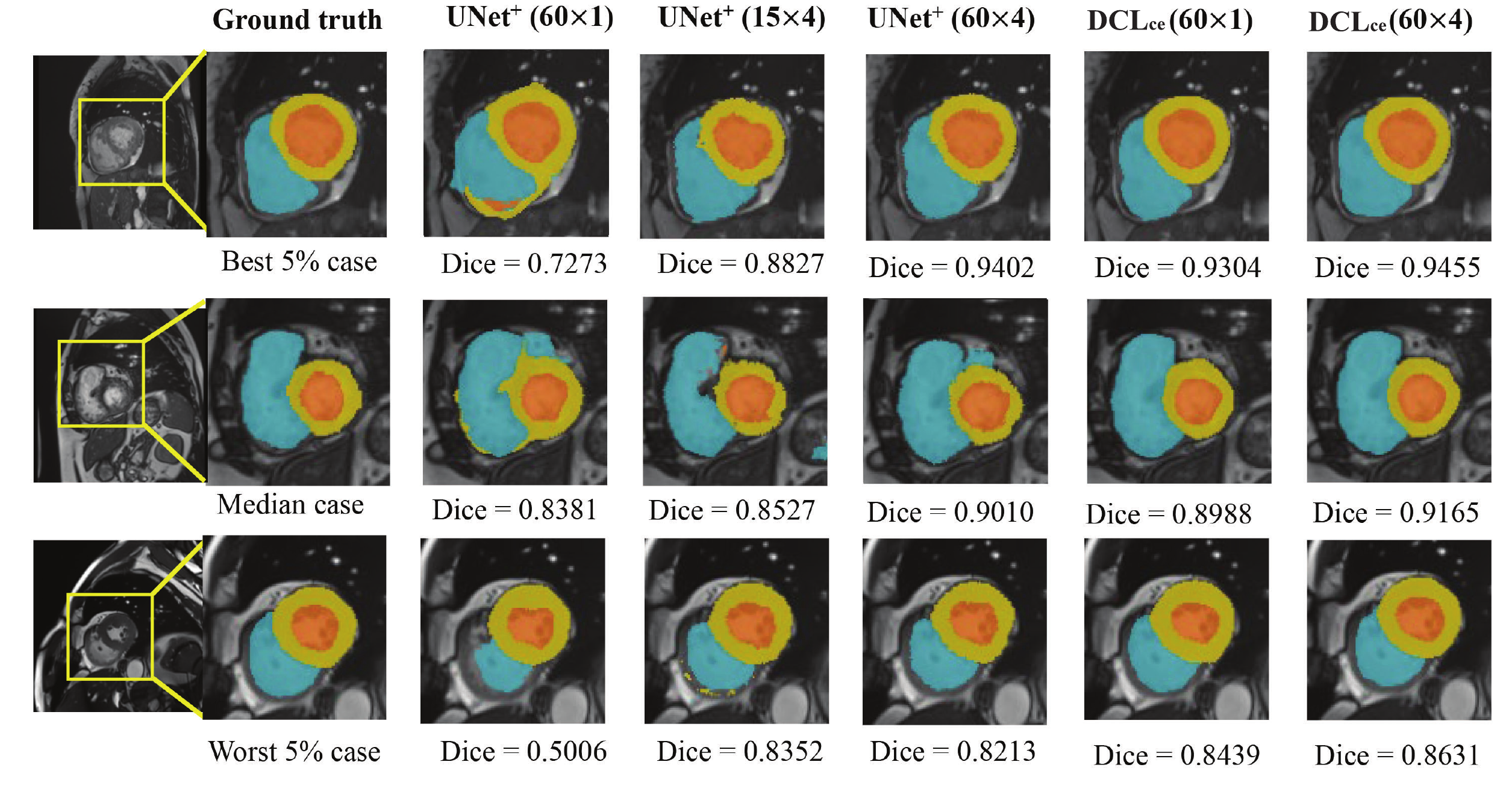}
		\caption{Quantitative comparison of the proposed method with partially- and fully- supervised baselines on ACDC dataset. Our approach DCL$_{ce}$(60$\times$1) achieves comparable results to fully-supervised baseline UNet$^+$(60$\times$4), utilizing only 1/4 training labels. The three subjects were the best 5\%, the median and the worst 5\% cases with regard to the Dice scores of the results of fully-supervised baseline UNet$^+$(60$\times$4).} 
		\label{fig7}
	\end{figure*}

	Table~\ref{tab6} presents the evaluation results of conventional loss functions and their DCL forms on the partially-supervised LGE CMR segmentation. 
	The results demonstrated that the compatibility framework achieved much better results on all the three loss functions.
	Particularly, DCL performed evidently better on MMEE, with an Dice score increase of $4.8\%$, \textit{i.e.}, $0.817$ vs $0.769$, indicating that DCL is generally applicable to existing loss functions and could provide additional supervision for missing labels. 
	Notably, with CE and CE+Dice, DCL surpassed the conventional methods by large margins of $32\%$ ($0.812$ vs $0.492$) and $23.9\%$ ($0.785$ vs $0.546$), respectively.
	One can also observe that DCL effectively reduced the performance difference brought by different loss functions, \textit{i.e.}, from $27.7\%$ ($0.769$ vs $0.492$) to $3.2\%$ ($0.817$ vs $0.785$).
	
	
	\subsection{Performance and comparisons with literature}
	\label{sec4.3}
	
	\subsubsection{ACDC Dataset}
	\begin{figure*}[thb]
		\centering
		\includegraphics[width=0.8\textwidth]{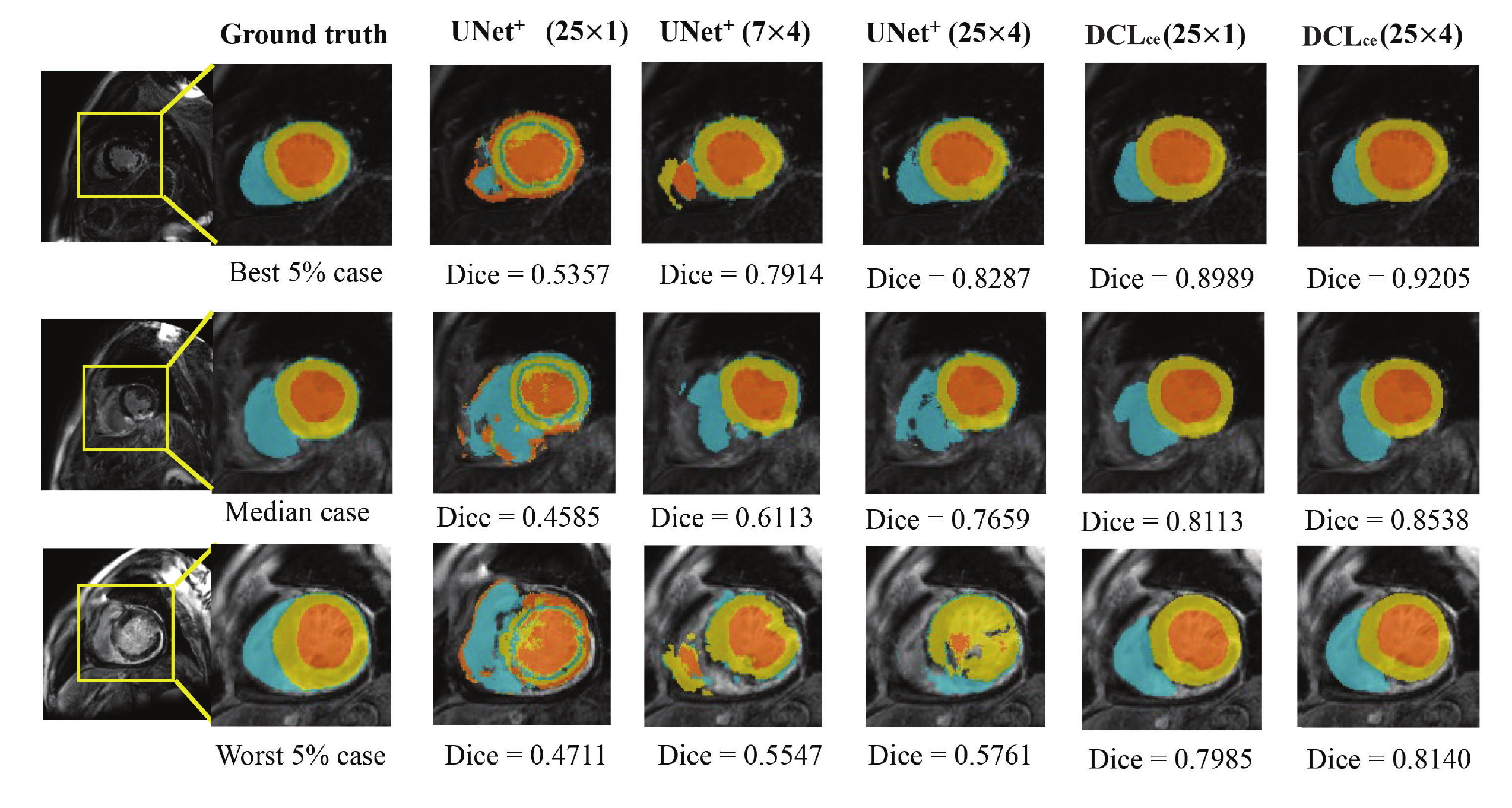}
		\caption{Quantitative comparison of the proposed method with partially- and fully- supervised baselines on MS-CMR dataset. Our approach DCL$_{ce}$(25$\times$1) produces improved results compared to the fully-supervised baseline UNet$^+$(25$\times$4), utilizing only 1/4 training labels.
			The three subjects were the best 5\%, the median and the worst 5\% cases with regard to the Dice scores of the results of fully-supervised baseline UNet$^+$(25$\times$4).} 
		\label{fig8}
	\end{figure*}
	
	For ACDC dataset, we referred to the result from \cite{bernard2018deep}, including a UNet trained on 100 fully annotated subjects and the range of agreement provided by ACDC challenge organizers.
	
	Table~\ref{tab2} presents the Dice scores on the ACDC dataset for different segmentation schemes.
	For partial annotation supervision,  one can observe that DCL provides substantial improvements over other fully-supervised and partially-supervised schemes, especially in the challenging RV segmentation of ES phase.
	Particularly, UNet$^+_F$(15$\times$4), with full supervision from 15 subjects, performed much better than UNet$^+_P$(60$\times$1), with partial supervision from 60 subjects, i.e. 0.834 vs. 0.778, indicating the ignorance of unlabeled structures reduces the performance of UNet.
	DCL$_{ce}$(60$\times$1) provided remarkable gains in performance compared with other CE-based schemes (UNet$^+_P$, UNet$^{+}_F$). 
	The partially-supervised method MMEE\cite{shi2020marginal} also achieved a substantial improvement in performance compared to UNet$^+_P$ and UNet$^{+}_F$.
	By applying DCL to MMEE, we see that DCL further improved the segmentation on the challenging structure ES-RV by $4.2\%$. 
	This shows the complementary benefits of the two methods.
	
	In the full annotation supervision, DCL(60$\times$4) performed comparably to the SOTA method UNet$^+$ with 100 training cases \cite{bernard2018deep}.
	For example, the Dice scores of DCL$_{ce}$(60$\times$4) surpassed the average Dice score (Avg)~\cite{bernard2018deep} trained on 100 subjects with full labels; and the Dice score of DCL$_{ce}$(60$\times$4) is even better than UNet$^+$(100$\times$4) by $0.6\%$ on ES-LV.
	Note that segmentation of ES-RV is a challenging task, where all the methods performed relatively poorly, indicating more training data and supervision are needed.
	Nevertheless, with the same training set DCL$_{ce}$(60$\times$4) achieved much better Dice on this challenging task, with an increase of 7.2\% (0.858 vs. 0.786), compared to UNet$^+$(60$\times$4), though the gains on easy tasks were marginal.

	Finally, \zkreffig{fig7} visualizes typical results using our implementation of compared methods on the worst 5\%, median and best 5\% cases of the fully-supervised UNet$^+$(60$\times$4). One can observe that the partially-supervised DCL$_{ce}$(60$\times$1) achieved comparable performance compared with the fully-supervised UNet$^+$(60$\times$4). The fully-supervised DCL$_{ce}$(60$\times$4) obtained the best results on all these cases.
	
	\begin{table}[htb]
		\caption{The performance (Dice scores) on LGE images from MSCMRseg dataset. $\text{Superscript}^{\dag}$ denotes the Dice was averaged after excluding one failed test case.}\label{tab3}
		\begin{center}
			\resizebox{1\linewidth}{!}{
				\begin{tabular}{lcccccc}
					\hline
					Methods&Total&Test& LV & MYO & RV &\multicolumn{1}{|c}{Avg}\ \\
					\hline
					\multicolumn{7}{l}{partial annotation supervision}\\
					\hline
					\multicolumn{1}{l|}{UNet$^+_P$}&\multicolumn{1}{c|}{25$\times$1}&\multicolumn{1}{c|}{15}&.658$\pm$.217&.556$\pm$.081&.262$\pm$.102&\multicolumn{1}{|c}{.492}\\
					\multicolumn{1}{l|}{UNet$^+_F$}&\multicolumn{1}{c|}{7$\times$4}&\multicolumn{1}{c|}{15}&.754$\pm$.097&.615$\pm$.090&.541$\pm$.159&\multicolumn{1}{|c}{.637}\\
					\multicolumn{1}{l|}{DCL$_{ce}$}&\multicolumn{1}{c|}{25$\times$1}&\multicolumn{1}{c|}{15}&.893$\pm$.048&.774$\pm$.049&$\bm{.769}\pm$.140&\multicolumn{1}{|c}{.812}\\
					\multicolumn{1}{l|}{MMEE\cite{shi2020marginal}}
					&\multicolumn{1}{c|}{25$\times$1}&\multicolumn{1}{c|}{15}&.889$\pm$.059&.750$\pm$.092&.667$\pm$.092&\multicolumn{1}{|c}{.769}\\
					\multicolumn{1}{l|}{DCL$_{\mathrm{MMEE}}$}&\multicolumn{1}{c|}{25$\times$1}&\multicolumn{1}{c|}{15}&$\bm{.897}\pm$.034&$\bm{.786}\pm$.050&$\bm{.769}\pm$.101&\multicolumn{1}{|c}{$\bm{.817}$}\\
					\hline
					\multicolumn{7}{l}{full annotation supervision}\\
					\hline
					\multicolumn{1}{l|}{UNet$^+_F$}&\multicolumn{1}{c|}{25$\times$4}&\multicolumn{1}{c|}{15}&.814$\pm$.120&.693$\pm$.008&.659$\pm$.110&\multicolumn{1}{|c}{.722}\\
					\multicolumn{1}{l|}{SRSCN\cite{yue2019cardiac}$^{\dag}$}&\multicolumn{1}{c|}{25$\times$4}&\multicolumn{1}{c|}{14}&.915$\pm$.052&.812$\pm$.105&.882$\pm$.084&\multicolumn{1}{|c}{.870}\\
					\multicolumn{1}{l|}{SRSCN\cite{yue2019cardiac}}&\multicolumn{1}{c|}{25$\times$4}&\multicolumn{1}{c|}{15}&.854&.758&.823&\multicolumn{1}{|c}{.812}\\
					\multicolumn{1}{l|}{DCL$_{ce}$}&\multicolumn{1}{c|}{25$\times$4}&\multicolumn{1}{c|}{15}&.913$\pm$.038&.811$\pm$.051&.810$\pm$.119&\multicolumn{1}{|c}{.845}\\
					\hline
			\end{tabular}}
		\end{center}
	\end{table}

	\begin{table*}[tbh]
		\caption{The performance (Dice scores) on MMWHS dataset for partially- and fully- supervised segmentation.  Column \textit{Total} indicates the total number of annotated labels in training images.}
		\scriptsize
		\begin{center}
			\resizebox{1\textwidth}{!}{
				\begin{tabular}{lccccccccc}
					\hline
					Methods&\multicolumn{1}{|c|}{Total} & LV & MYO & RV & LA & RA & AO & PA & \multicolumn{1}{|c}{WH}\\
					\hline
					\multicolumn{10}{l}{partial annotation supervision}\\
					\hline
					\multicolumn{1}{l|}{nnUNet$_P$}&\multicolumn{1}{c|}{20 $\times$ 2}&.864$\pm$.209&.743$\pm$ .210&.772$\pm$.176&.822$\pm$.093&.805$\pm$.112&.821$\pm$.150&.587$\pm$.269&\multicolumn{1}{|c}{.803$\pm$.123}\\
					\multicolumn{1}{l|}{nnUNet$_F$}&\multicolumn{1}{c|}{5 $\times$ 8}&.879$\pm$.138&.775$\pm$.177&.831$\pm$.180&$\bm{.834}\pm$.120&.856$\pm$ .088&$\bm{.845}\pm$.167&$\bm{.754}\pm$.166&\multicolumn{1}{|c}{.840$\pm$.135}\\
					\multicolumn{1}{l|}{DCL$_{ce}$}&\multicolumn{1}{c|}{20 $\times$ 2} &$\bm{.893}\pm$.090 &$\bm{.800}\pm$.088 &$\bm{.876}\pm$.103 &.787$\pm$.134 &$\bm{.867}\pm$.083&.795$\pm$.157&.714$\pm$.177&\multicolumn{1}{|c}{$\bm{.857}\pm$.061}\\
					\hline
					\multicolumn{10}{l}{full annotation supervision}\\
					\hline
					\multicolumn{1}{l|}{Avg\cite{zhuang2019evaluation}}&\multicolumn{1}{c|}{20 $\times$ 8}&.861$\pm$.151&.742$\pm$.140&.824$\pm$.144&.815$\pm$.131&.826$\pm$.135&.797$\pm$.159&.705$\pm$.192&\multicolumn{1}{|c}{.820$\pm$.104}\\
					\multicolumn{1}{l|}{GUT(DL)\cite{payer2017multi}}&\multicolumn{1}{c|}{20 $\times$ 8}&.916$\pm$.043&.778$\pm$.088&.868$\pm$.094&.855$\pm$.050&.881$\pm$.037&.838$\pm$.048&.731$\pm$.187&\multicolumn{1}{|c}{.863$\pm$.043}\\
					\multicolumn{1}{l|}{UOL(MAS)\cite{heinrich2017mri}}&\multicolumn{1}{c|}{20 $\times$ 8}&.918$\pm$.040&.781$\pm$.076&.871$\pm$.070&$\bm{.886}\pm$.050&.873$\pm$.046&.878$\pm$.068&.804$\pm$.106&\multicolumn{1}{|c}{.870$\pm$.035}\\
					\multicolumn{1}{l|}{DCL$_{ce}$}&\multicolumn{1}{c|}{20 $\times$ 8}&$\bm{.919}\pm$.039&$\bm{.840}\pm$.050&$\bm{.903}\pm$.061&.847$\pm$.070&$\bm{.884}\pm$.037&$\bm{.880}\pm$.063&$\bm{.807}\pm$.107&\multicolumn{1}{|c}{$\bm{.887}\pm$.032}\\
					\hline
			\end{tabular}}
		\end{center} 
		\label{tab4}\end{table*}
	
	\begin{figure*}[thb]
		\centering
		\includegraphics[width=0.95\textwidth]{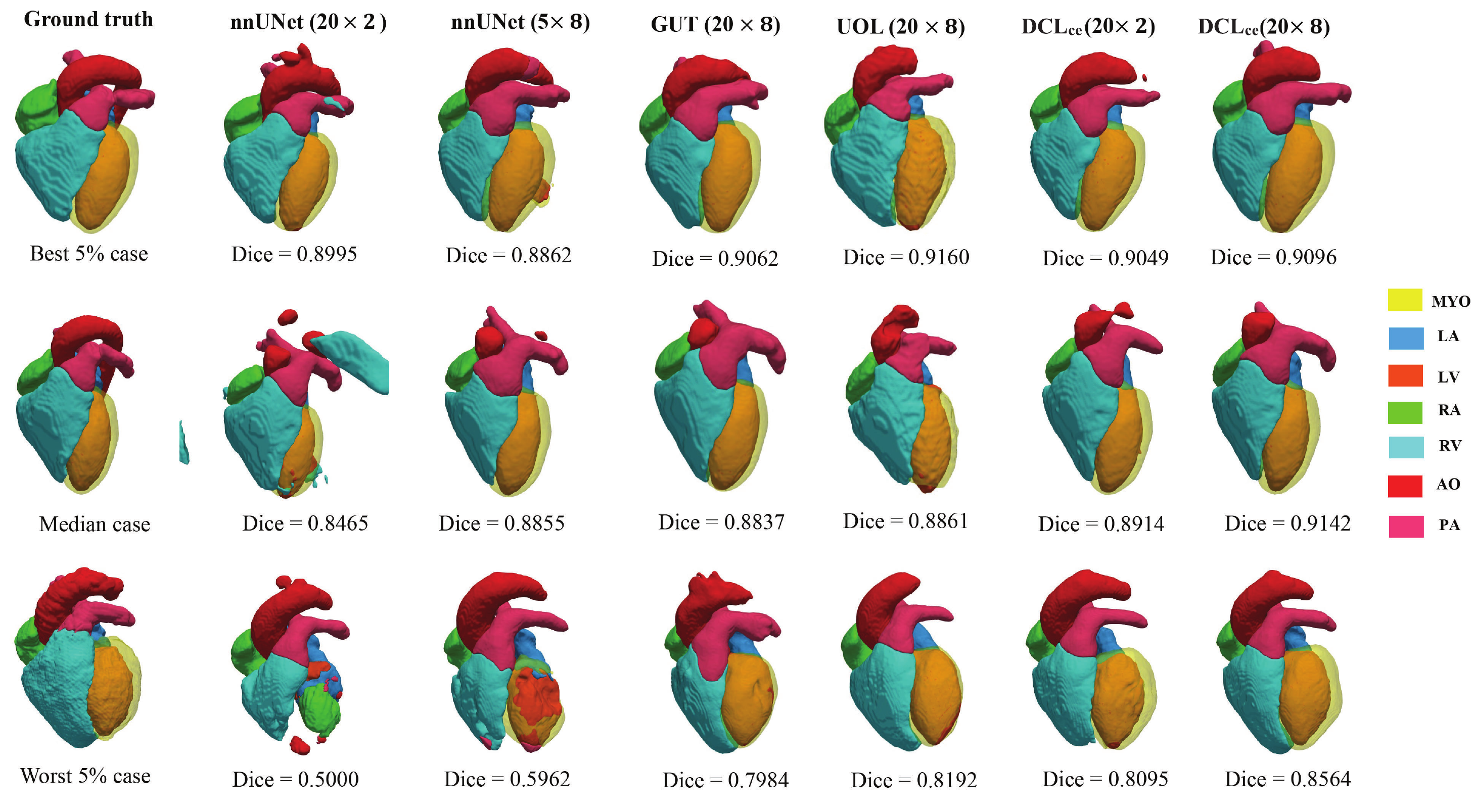}
		\caption{Quantitative comparison of the proposed method on MMWHS dataset.  The three subjects were the best 5\%, the median and the worst 5\% cases with regard to the Dice scores of the results of average Dice scores Avg(20$\times$8) provided by challenge organizers.} 
		\label{fig9}
	\end{figure*}
	
	\subsubsection{MSCMR Dataset}
	For LGE CMR images from MSCMRseg,  besides our implementation of UNet and partially-supervised segmentation method MMEE, we further referred to two results reported in \cite{yue2019cardiac}, i.e. (1) the accuracy of the state-of-the-art (SOTA) method SRSCN, which is composed of a shape reconstruction neural network and a spatial constraint network on 14 test images, (2) SRSCN performance on 15 test images.
	To show the effect of DCL more obviously, we conducted additional seven sets of experiments on the MSCMRseg dataset with less training data.
	
	Table~\ref{tab3} presents the results on the LGE CMR segmentation from MSCMRseg dataset. Note that this is a much more challenging task compared to the ACDC segmentation, as LGE CMR segmentation \textit{per se} is more complex and the training set is smaller.
	The results show that DCL$_{ce}$(25$\times$1) with partial labels not only performed much better than UNet$^+_P$(25$\times$1) and UNet$^+_F$(7$\times$4),
	but also achieved better or no worse mean Dice than all the SOTA results obtained by the fully-supervised segmentation using the same number of training and test images.
	Here, the fully supervised segmentation results include the mean Dice scores of 14 test cases from Yue \textit{et al} \cite{yue2019cardiac}, where one failed case was excluded for calculation.
	Yue \textit{et al} failed one case when they evaluated SRSCN and UNet on the test set. For fair comparisons, the reported results provided by authors (excluding the failure case) and adjusted results (including the failure case) are both listed in Table~\ref{tab3}. 
	The results adjusted by failure case are denoted by $\text{superscript}^{\dag}$. DCL$_{ce}$ achieved a performance increase of $12.2\%$ over implemented UNet baseline and $3.3\%$ over SRSCN.
	
	We see that the marginal (target adaptive) loss and exclusive loss (MMEE) provide a marginal improvement when combined with DCL. This further demonstrates the effectiveness of proposed DCL and the complementary advantages of the two approaches.

	\zkreffig{fig8} provides visualizations of typical segmentation results on the worst 5\%, median, and best 5\% cases of fully-supervised UNet$^+$(25$\times$4). DCL$_{ce}$ achieved the best results on all these cases, compared to the fully- and partially-supervised baselines. Note that with only $1/4$ labels, DCL$_{ce}$(25$\times$ 1) obtained much better performance than supervised UNet$^+$(25 $\times$ 4) on all these cases.

	\subsubsection{MMWHS Dataset}
	For MMWHS dataset, we referred to the experiment results reported in \cite{zhuang2019evaluation}, including state-of-the-art deep learning method GUT\cite{payer2017multi} trained on 20 fully annotated subjects and the average performance provided by MMWHS challenge organizers. 
	
	Table~\ref{tab4} shows the segmentation performance on the MMWHS dataset, with partial supervision and full supervision. For partially-supervised learning, our method achieved a performance increase of $5.4\%$ over the baseline of nnUNet$_P$(20$\times$2) on whole heart (WH) segmentation by utilizing the compatible learning. Notably, the approach worked well on the difficult RV label, improving the dice score of $10.4\%$, \textit{i.e.}, $0.876$ vs $0.772$, compared to nnUNet$_P$(20$\times$2). With $1/4$ labels, DCL$_{ce}$ also surpassed the average result of fully supervised segmentation provided by the MMWHS segmentation challenge organizers~\cite{zhuang2019evaluation}, by a large margin of $3.7\%$, \textit{i.e.}, $0.857$ vs $0.820$. 
	
	For fully-supervised segmentation, we used two state-of-the-art methods as baselines, \textit{i.e}, GUT\cite{payer2017multi} and UOL\cite{heinrich2017mri}. 
	GUT and UOL are based on deep learning (DL) and multi-atlas segmentation (MAS), respectively. They have achieved the first and second best Dice in the MMWHS challenge. 
	Compared to the DL-based SOTA method, the proposed DCL$_{ce}$ outperformed GUT by $2.4\%$ on WH, \textit{i.e.}, $0.887$ vs $0.863$. The performance gain was even more significant on the difficult RV label, with an increase of $7.9\%$. 
	DCL$_{ce}$ also obtained an improvement of $1.7\%$ Dice score on WH, compared to MAS-based UOL ($0.887$ vs $0.870$), setting the new state-of-the-art performance.
	\zkreffig{fig9} provides visualization of typical results on the worst 5\%, median and best 5\% cases of the average performance reported in \cite{zhuang2019evaluation}.
	One can see that DCL$_{ce}$ can achieve the best performance with partial and full annotations.
	DCL$_{ce}$(20$\times$2) under the same partial supervision achieved better performance than nnUNet(20$\times$2) and nnUNet(5$\times$8), and was comparable to the performance of fully supervised UCL(20$\times$8).
	Moreover, compared to GUT(20$\times$8) and UOL(20$\times$8), DCL$_{ce}$(20$\times$8) performed significantly better, especially on MYO in the worst 5\% case.
	This verified the effectiveness and advantages of the proposed compatible learning.
	
	\section{Conclusion} \label{section6}
	In this paper, we have presented the deep compatible learning (DCL) framework, which formulates partially-supervised segmentation task as an optimization problem compatible with missing labels. 
	The proposed framework jointly incorporates conditional compatibility and dual compatibility to provide substantial supervision for unlabeled structures. 
	Our framework is generally applicable to the existing loss functions that are compatible with fully-supervised learning.
	Results showed that the deep compatible framework, with only partial labels, could endow conventional loss functions with significant improvement and achieve comparable performance to the fully-supervised methods.
	
	\ifCLASSOPTIONcompsoc
	\section*{Acknowledgments}
	\else
	\section*{Acknowledgment}
	\fi 
	This work was funded by the National Natural Science Foundation of China (grant no. 61971142, 62111530195 and 62011540404) and the development fund for Shanghai talents (no. 2020015)
	The authors would like to thank Fuping Wu, Shangqi Gao, Zheyao Gao, Bomin Wang, Junyi Qiu, Hangqi Zhou for useful comments and proof read of the manuscript.

	\ifCLASSOPTIONcaptionsoff
	\newpage
	\fi

	\bibliographystyle{IEEEtran}
	\normalem
	\bibliography{IEEEabrv,citation}
	%
	%
	%
	
	%
	
	\begin{IEEEbiography}[{\includegraphics[width=1in,height=1.25in,clip,keepaspectratio]{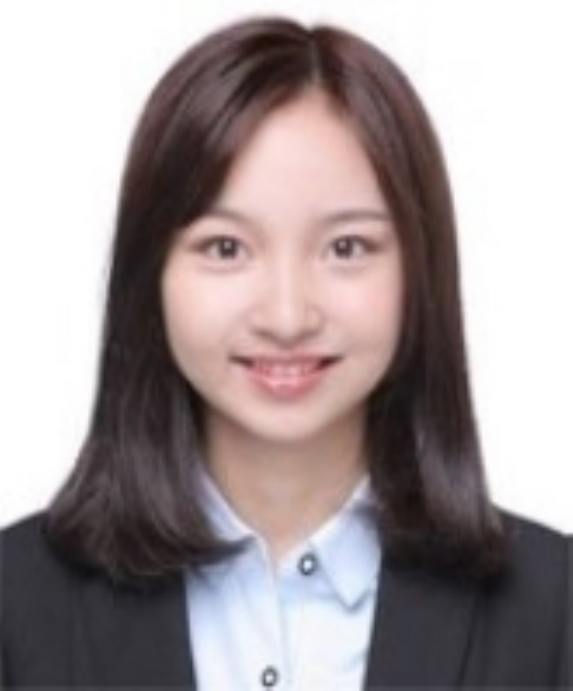}}]{Ke Zhang} 
		is a post graduate student at School of Data Science, Fudan University, Shanghai, China, supervised by Prof. Xiahai Zhuang. She obtained her bachelor degree in statistics from Fudan University. Her research interests include computer vision, medical image analysis and AI in healthcare.
	\end{IEEEbiography}
	
	\begin{IEEEbiography}[{\includegraphics[width=1in,height=1.25in,clip,keepaspectratio]{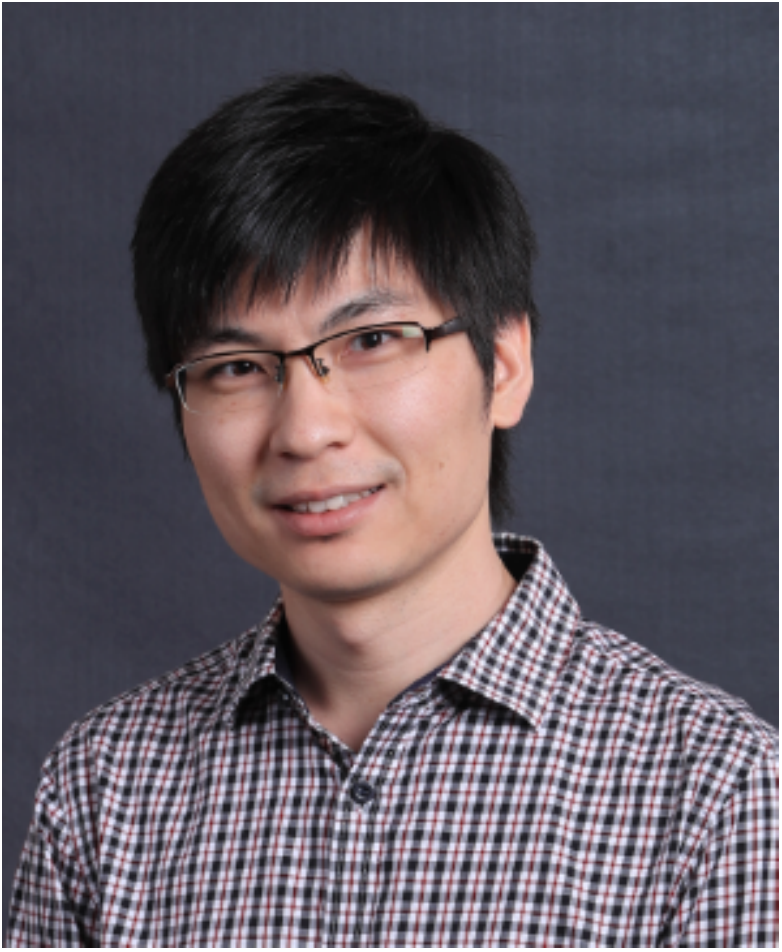}}]{Xiahai Zhuang}
		is a professor with the School of Data Science, Fudan University.
		He graduated from the department of computer science, Tianjin University, received the MS degree in computer science from Shanghai Jiao Tong University, and the doctorate degree from University College London.  His research interests include medical image analysis, image processing, and computer vision. His works have been nominated twice for the MICCAI Young Scientist Awards (2008, 2012).
	\end{IEEEbiography}
	
	

	
	

\end{document}